\documentclass{article}

\usepackage{arxiv}

\usepackage[utf8]{inputenc} 
\usepackage[T1]{fontenc}    
\usepackage{hyperref}       
\usepackage{url}            
\usepackage{booktabs}       
\usepackage{amsfonts}       
\usepackage{amsmath}
\usepackage{nicefrac}       
\usepackage{microtype}      
\usepackage{lipsum}		
\usepackage{graphicx}
\usepackage{natbib}
\usepackage{doi}

\usepackage{tabularx}
\usepackage{subcaption}
\usepackage{booktabs, multirow}
\usepackage{float}
\usepackage{marginnote}
\usepackage{tabularx}
\usepackage{fancyhdr}
\usepackage{rotating}
\usepackage{adjustbox}
\usepackage{microtype}
\usepackage{algorithm}
\usepackage{algpseudocode}
\usepackage{subcaption}
\usepackage{booktabs, multirow}
\usepackage{marginnote}
\usepackage{tabularx}
\usepackage{colortbl}
\usepackage{graphicx}%
\usepackage{multirow}%
\usepackage{amsmath,amssymb,amsfonts}%
\usepackage{amsthm}%
\usepackage{mathrsfs}%
\usepackage[title]{appendix}%
\usepackage[dvipsnames]{xcolor}%
\usepackage{textcomp}%
\usepackage{manyfoot}%
\usepackage{booktabs}%
\usepackage{algorithm}%
\usepackage{algorithmicx}%
\usepackage{algpseudocode}%
\usepackage{listings}%
\usepackage{amsmath}
\usepackage{svg}

\title{X-SHIELD: Regularization for eXplainable Artificial Intelligence}

\author{ \href{https://orcid.org/0000-0002-5029-9106}{\includegraphics[scale=0.06]{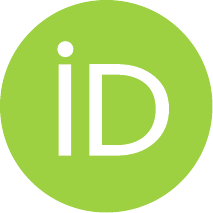}\hspace{1mm}Iván Sevillano-García} \\
	Department of Computer Science and \\
    Artificial Intelligence\\
	Andalusian Research Institute in Data Science and \\Computational Intelligence (DaSCI)\\
	University of Granada, Granada, 18071 \\
	\texttt{isevillano@go.ugr.es} \\
	\And
	\href{https://orcid.org/0000-0003-3952-3629}{\includegraphics[scale=0.06]{orcid.pdf}\hspace{1mm}Julián Luengo} \\
	Department of Computer Science and\\
    Artificial Intelligence\\
	Andalusian Research Institute in Data Science and \\Computational Intelligence (DaSCI)\\
	University of Granada, Granada, 18071 \\
	\texttt{julianlm@decsai.ugr.es} \\
	  \AND
	\href{https://orcid.org/0000-0002-7283-312X}{\includegraphics[scale=0.06]{orcid.pdf}\hspace{1mm}Francisco Herrera} \\
	Department of Computer Science and\\
    Artificial Intelligence\\
	Andalusian Research Institute in Data Science and \\Computational Intelligence (DaSCI)\\
	University of Granada, Granada, 18071 \\
	\texttt{herrera@decsai.ugr.es} \\
}

\begin{document}
\maketitle

\begin{abstract}
As artificial intelligence systems become integral across domains, the demand for explainability grows, the called eXplainable artificial intelligence (XAI).
Existing efforts primarily focus on generating and evaluating explanations for black-box models while a critical gap in directly enhancing models remains through these evaluations. 
It is important to consider the potential of this explanation process to improve model quality with a feedback on training as well. XAI may be used to improve model performance while boosting its explainability.
Under this view, this paper introduces Transformation - Selective Hidden Input Evaluation for Learning Dynamics~(T-SHIELD), a regularization family designed to improve model quality by hiding features of input, forcing the model to generalize without those features. Within this family, we propose the XAI - SHIELD~(X-SHIELD), a regularization for explainable artificial intelligence, which uses explanations to select specific features to hide.
In contrast to conventional approaches, X-SHIELD regularization seamlessly integrates into the objective function enhancing model explainability while also improving performance. 
Experimental validation on benchmark datasets underscores X-SHIELD's effectiveness in improving performance and overall explainability. 
The improvement is validated through experiments comparing models with and without the X-SHIELD regularization, with further analysis exploring the rationale behind its design choices.
This establishes X-SHIELD regularization as a promising pathway for developing reliable artificial intelligence regularization.

\end{abstract}


\section{Introduction}

Artificial Intelligence (AI) has made extraordinary strides, transforming how we approach real-world challenges. 
These advances lead to the parallel development of needs such as explainability, which gives rise to the area called explainable AI (XAI). A general definition of XAI is provided in~\citep{arrieta2020explainable} as "given an audience, an explainable Artificial Intelligence is one that produces details or reasons to make its functioning clear or easy to understand".

Recognizing the need for explainability,  researchers have introduced a plethora of XAI techniques,
such as the use of counterfactuals~\citep{wachter2017counterfactual}, 
generation of explanations based on feature importance, such as LIME~\citep{ribeiro2016should} or SHAP~\citep{lundberg_unified_2017} in a model agnostic way, or such as DeepLIFT~\citep{shrikumar2017learning} or Integrated Gradients~\citep{sundararajan2017axiomatic} in a model dependant way. 
These techniques aim to generate explanations that shed light on the decision-making processes, facilitating human understanding. 
For a comprehensive analysis of XAI techniques, we direct the reader to the following overview~\citep{bennetot2024practical}. On the other hand, the field has witnessed the proposal of different kind of studies aimed at unraveling the complexities of black-box models (see the overview~\citep{ALI2023101805}). 
These approaches not only strive to make these models comprehensible to humans but also underscore the risks associated with neglecting the XAI perspective~\citep{kong2024toward}.  Recently,  new challenges and roadmaps for the development of the so called XAI 2.0 has been published~\citep{longo2024explainable}.

Focusing on the audiencies and purposes, in~\citep{biecekposition} is introduced a double perspective: Red XAI~(Research, Evaluate and Develop) and Blue XAI~(responsibLe, Legal, trUst and Ethics). Red XAI  uses explainability to improve AI from the developer's point of view while Blue XAI focuses on the on promoting explainability for users. The Red XAI is an approximation of the importance of XAI in the design of AI models with different purposes. XAI may be used to improve model performance while boosting its explainability.

It is at this juncture where we can observe the introduction of various methods to obtain explanations for models but not a clear methodology to choose between those methods of obtaining explanations. 
Different ways of evaluating these explanations, both qualitatively and quantitatively, have been proposed~\citep{amparore2021trust}. 
We highlight REVEL~\citep{sevillano2023revel}, a consistent and mathematically robust framework for quantitatively measuring different aspects of the explanations generated by Local Linear Explanations (LLEs). However, this knowledge measured from the explanations has not been used until now to improve model performance while boosting its explainability.

That is why, once the quality metrics of explanations have been defined, we should be able to introduce this quality criterion into the learning models. This process involves systematically incorporating a quality bias within the models themselves. By doing so, we ensure that the learning algorithms prioritize not just accuracy or performance, but also the clarity and explainability of their outputs, and user-friendly as final goal.  This could involve refining the training processes, adjusting model parameters, or even redesigning model architectures to inherently favor explanations that meet the defined quality metrics.

Regularization techniques have emerged as a common solution. 
By incorporating an additional terms into the objective function, these techniques ensure that models meet specific desired criteria. 
Weight-decay, dropout, data augmentation, early stopping, and more examples of regularizations are described in~\citep{moradi2020survey}. 
From this perspective, if it is possible to quantify explanations quality, we might improve the model with this insights. 
Based on this approach, on~\citep{weber2023beyond} overview and classify different thoretical approaches to improve the quality of XAI-based models.

In response to the discussed challenges and opportunities presented by existing techniques we focus in this avenue to improve model performance while boosting its explainability. we propose the Transformation - Selective Hidden Input Evaluation for Learning Dynamics~(T-SHIELD) regularization family. 
This family is developed to enhance model quality by hiding features from the input data forcing the model to learn with fewer features. 
Within this family, we concretize the proposal with eXplainable artificial intelligence-SHIELD~(X-SHIELD), which thanks to the XAI perspective, selects the least important features to hide.
The theoretical foundations of T-SHIELD and X-SHIELD are explored in detail, outlining how the selection of input features can serve to improve both model performance and explainability. This aligns it with the Red XAI perspective as a XAI-based model.

We conduct a comprehensive experimental analysis to test whether the model is enhanced by X-SHIELD regularization approach with particular emphasis on the gain in explainability provided by the proposal.
This evaluation provides insights into the efficacy of X-SHIELD regularization offering a practical perspective on its utility.

This paper unfolds as follows: 
Section \ref{sec:background} establishes the background, emphasizing the ability to escalate the demand for explaining AI systems and introducing how regularizations improve certain quality criteria in models in AI. 
On Section \ref{sec:proposal}, we unveil the T-SHIELD regularization family and the X-SHIELD regularization, offering preliminary definitions and intuition of the proposal. 
Section \ref{sec:experimental_setup} outlines the experimental setup, incorporating benchmark datasets and implementation details. 
We present the experimental findings on Section \ref{sec:experimental_results}, assessing the effectiveness of the regularization in performance and the importance of the selection of the transformation T no enhance model explainability. 
Finally, Section \ref{sec:conclusions} concludes the paper, summarizing contributions, highlighting the importance of reliability in AI, and charting potential avenues for future research.

\section{Regularization in XAI}
\label{sec:background}
In this section, we detail different theoretical concepts that have influenced the development of the proposed X-SHIELD regularization. 
In particular, we firstly describes what constitutes a feature in XAI, the difference between white-box and black-box perspectives, and what a regularization theoretically consists of. We finally describe the XAI metrics used in the experimental results comparison.

In Section~\ref{sec:xai}, we introduce different concepts and definitions of the scope of XAI are , which are necessary for the development of this work.
In Section~\ref{sec:xAI_Ev}, we show the current development of the study of approaches to evaluate explanations. 
Finally, in Section~\ref{sec:reg}, we review the different ways of guiding or regularizing training to impose different criteria on the final models.

\subsection{XAI Concepts and definitions}
\label{sec:xai}
Typically, humans hesitate to embrace methods that lack clear interpretation, manageability, and trustworthiness, especially with the growing call for ethical AI. In these systems, prioritizing performance alone leads to less transparent systems. Indeed, there is a recognized balance to be struck between a model's effectiveness and its transparency~\citep{arrieta2020explainable}.
However, an improvement in the understanding of any system may help to correct its deficiencies.
It is in this context where XAI arises, enabling humans to understand and trust their AI companion.

As indicated in~\citep{arrieta2020explainable}, ``\emph{given an audience, an explainable Artificial Intelligence produces details or reasons to make its functioning clear or easy to understand}''. 
Derived from this concept, an \emph{explanation} is an interface between humans and an AI decision-maker that is both comprehensible to humans and an accurate proxy of AI~\citep{arrieta2020explainable}.
However, in the discourse on XAI, a notable challenge is the absence of a universal consensus on the definitions of many key terms~\citep{longo2024explainable}. 
Thus, the landscape of XAI research is characterized by the presence of diverse taxonomies that aim to define and categorize various approaches to explanations~\citep{angelov2021explainable}.

XAI taxonomies agree that, at a high level, explanations in the context of XAI can be broadly categorized into agnostic explanations and model-dependent explanations, depending on the use of the information that is available or that we want to use from the AI model. 
Recently, in~\citep{bodria2023benchmarking}, the authors account for the most representative explainers categorized by data types and introduce the most well-known and used explanation to date.
They also gather the fundamental distinctions used to annotate the explainers used in previous taxonomies: intrinsic, post-hoc, global, local, model-agnostic, and model-specific techniques.

Model-agnostic explanations~\citep{ribeiro2016should} refer to XAI methods that aim to provide insights into model predictions without relying on the internal structures of the model. 
This perspective eliminates the need for any special features of the model, such as using well-known layers or even knowing whether the internal model is differentiable, that is, there is no bias associated with the structure itself. 
Furthermore, it can be applied to any current or future model.
The input and output are the only information that influence this explanation, making model-agnostic explanations ideal for auditing models.
However, since it lacks connection to the internal structure of the model, these explanations can't be used to enhance the model in any aspect through optimization methods such as gradient descent. It is also a disadvantage that, to compute these explanations, it is necessary to evaluate the black-box model a large number of times to obtain a reliable explanation.

Model-specific explanations like GradCam~\citep{selvaraju2017grad} or derivatives, on the other hand, rely on understanding the internal structures of the model to provide insights into its predictions.
This perspective implies a requirement for specific model knowledge, such as the specific layers of the model or determining the differentiability. 
It's also important to consider that these explanations are faster than model-agnostic approaches since there's no need to evaluate the model multiple times.
The connection to the internal structure allows for potential model enhancements through optimization methods like gradient descent.
However, it could introduce biases associated with the model's structure~(i.e., Will the explanation change if we change the layer where we apply GradCam?).
This makes dependant explanations valuable for improving models but may raise concerns in auditing the model.

One of the major distinctions highlighted in~\citep{mcdermid2021artificial} among types of explanations, regardless of whether they are model-agnostic or model-dependent, is the differentiation between example-based and feature importance-based explanations.

Example-based explanations~\citep{wachter2017counterfactual} typically explain by comparing the examples to be explained with other examples the model has seen before by proximity in the feature space. This type of explanation is generated when the model finds examples within its dataset that are similar to the one to be explained, either same-class examples, showing the similarities betweeen them, or class-changed examples, also called counterfactuals, enhancing the main differences that makes the model change its decision.

Feature importance explanations are explanations where each feature has an associated importance concerning each of the model outputs. 
In model-agnostic explanations, this associated importance matrix comes from a linear regression that approximates the derivative of the model, which gives the name of Local Linear Explanations~(LLE), like LIME~\citep{ribeiro2016should} or SHAP~\citep{lundberg_unified_2017}. 
In model-dependent explanations, this importance is derived through calculations involving derivatives or similar methods, such as saliency maps~\citep{tomsett2020sanity}. 

This work is developed within the feature importance explanations so we shall formalize the terms used.
Let $f:X \xrightarrow{}Y$ a black-box model to be explained, where $X\subset \mathbb{R}^n$ and $Y\subset \mathbb{R}^m$.
Then, an explanation is a matrix 
$\mathcal{A}\subset \mathbb{R}^{nm}$ where 
each $a_{i,j}$ is the importance of the feature $i$ for the output $j$. 
In the case of linear models, let $g:X \xrightarrow{}Y$ be a linear model where $g(x) = Ax+b$. 
Then, the matrix $A$ is an explanation of the model g, that is, $\mathcal{A} = \dfrac{\delta g}{\delta X}(x) = A$. To obtain explanations of more complex models, a linear approximation is usually estimated in the vicinity of the example from which to obtain the final explanation~\citep{ribeiro2016should,lundberg_unified_2017} which is very similar to the conception of the derivative of saliency maps: $\mathcal{A} \sim \dfrac{\delta f}{\delta X}(x)$. 

Although the definition of feature can be approximated directly on the original space $X$, several approaches use a transformation on a feature space $F$, from which it is more intuitive and operative to create an explanation. 
In images, there is considered a region of pixels, called super-pixels, instead of pixels itself~\citep{ribeiro2016should,rehman2019dlime}.

\subsection{Evaluation of explanations in XAI}
\label{sec:xAI_Ev}


Evaluating the explanations can be approached from a qualitative or quantitative viewpoint, but no objective evaluation measures are known to select the best explainer to date~\citep{bodria2023benchmarking}.
A major current problem of explainability is the fact that a standard method to evaluate the different explanations is not available, that is, to choose which one is the best and in what sense.
There is a wide debate on how to evaluate the quality of the explanation methods, placing attention on their usefulness~\citep{bodria2023benchmarking}. 
It is important for an explanation to have a qualitatively meaningful explanation~\citep{doshi2017towards} but, to be able to compare objectively between explanations, there must be a quantitative measure to support the decision.
In~\citep{rosenfeld2021better} a set of quantitative metrics is proposed to measure the quality of explanations. However, they are specialized in rules-based explanations. 
In~\citep{sevillano2023revel}, the REVEL framework is proposed, with also five different quantitative metrics to evaluate agnostically different explanation metrics, independent of the explanation generation method.

\begin{table}[http]
    \scriptsize
    
\begin{tabularx}{\textwidth}{lX}\toprule
    \textbf{Name} &\textbf{What is evaluated} \\\cmidrule{1-2}
    \textbf{Local Concordance} &How similar is the explanation to the original black-box model on the original example \\\midrule
    \textbf{Local Fidelity} &How similar is the explanation to the original black-box model on a neighborhood of original example \\\midrule
    \textbf{Prescriptivity} &How similar is the explanation to the original black-box model on the closest neighbour that changes the class of the original example \\\midrule
    \textbf{Conciseness} &How brief and direct is the explanation \\\midrule
    \textbf{Robustness} &How much two explanations generated by the same explanation generator differ \\
    \bottomrule
\end{tabularx}
    
    \caption{Summary of the metrics developed~\citep{sevillano2023revel} and the qualitative aspect they measure}
    \label{tab:metrics_summary}
\end{table}

In Table~\ref{tab:metrics_summary}, we present a description of the qualitative aspect measured by the five metrics proposed in~\citep{sevillano2023revel}. 
Although there are several other proposals for metrics that can be found~\citep{10297629}, the metrics described here represent several important areas of explanation measurement considered in~\citep{rosenfeld2021better}:

\begin{itemize}
    \item \emph{Does the explanation behave like the model in the example to be explained?} Local Fidelity and Local Concordance measure this behavior. 
    \item \emph{Can this explanation be extrapolated far from the example to be explained?} Prescriptivity measures if a synthetic example constructed to differ from the original example ends up modifying the behavior of the model.
    \item \emph{How much can an explanation vary if it is generated several times?} Robustness measures how much two explanations of the same example differ if the method of generating explanations is stochastic. 
    \item \emph{The number of important features in the exlanation}. Conciseness measures the percentage of features with high involvement in the decisions of the model, being a very concise model when few features have high involvement and will be not very concise when there are many features with distributed influence in the decision of the model.
\end{itemize}

The metrics shown are quantitative and, when applied to model-dependent explanations, can be useful information to guide the models in the training stage to obtain better performance and explainability behavior.

\subsection{Towards the Integration of XAI via Regularization Methods}
\label{sec:reg}

Regularization techniques are widely used methods to impose quality criteria on different machine learning models~\citep{moradi2020survey}.
Most machine learning libraries already have regularization terms implemented as default~\citep{NEURIPS2019_9015}. 
Weight decay, dropout, normalization, or data augmentation are different techniques used to impose the quality criteria~\citep{zhang2024implicit}.

Some regularizations involve adding a term to the cost function, which, when minimized, helps optimize the quality criterion. L1, L2 or entrophy regularizations are example of regularizations that adds constraints explicitly on the loss function to impose some desired criteria. 
Other approaches introduce noise on the data to force the model to generalize better~\citep{kimanius2024data}. 
There are other works that analyse the mathematical formalisation of the regularizations in order to propose new regularizations that improve the model~\citep{DEMOL2009201}. 
It is worth noting that all these techniques cooperate, thus the value of each of them is simply determined by whether to use or not that particular technique, rather than by comparison with other regularization techniques.

Formally, for a regression problem with data set $X, Y$ and a parametric function $f_\Theta$ to optimize, the regularization is added to the cost function as follows:
$$
    cost\_function(\Theta,X,Y) = loss(f_\theta(X),Y) + regularization(\Theta),
$$
This cost function can be generalized by adding the input to the regularization as follows:
$$
    cost\_function(\Theta,X,Y) = loss(f_\theta(X),Y) + regularization(X,\Theta),
$$

This generalization manages to incorporate into the regularization term constraints that we want the model to have regarding the specific dataset, without allowing these constraints to bias the final prediction. 
In \citep{weber2023beyond}, different ways of improving models based on explanations are theoretically defined and classified depending on the part of the machine learning algorithm is used to improve. 
Within this classification of this work, the addition of regularization terms to the loss functions is categorized as a Loss Augmentation approach to improve explainability. There are two different approaches, depending on the need of human-explanations.

On the one hand, there are works that require explanations in order to ensure that a decision has been made for the right reasons. These approaches have the disadvantage of relying on human-labelled explanations. The obtention of these explanations is difficult and time-consuming because it requires human labeling besides the classification labeling~\citep{ijcai2017p371}.

On the other hand, other works propose to improve explanations without prior labeled explanations. These approaches propose to improve the characteristics of the explanations, without knowing whether these explanations are the ones that a human would provide, such as Saliency Guided~\citep{ismail2021improving} and SaliencyMix~\citep{uddin2020saliencymix}. These approaches do not require prior labeling. However, these approaches use specific explanations of the assigned class.
Moreover, the analysis of those methods are measured exclusively with the module of gradients as a measure of quality for explainability. To make sure of the improvement in explainability, it is mandatory to use quantitative descriptive measures of the resulting explanations.

\section{T-SHIELD regularization family and the X-SHIELD proposal}

\label{sec:proposal}

In this Section, we present the formalization of the X-SHIELD proposal. First, in Section~\ref{sec:tshield}, we define T-SHIELD, a regularization family to improve model quality by hiding specific input features. Second, in Section~\ref{sec:xshield_transformation}, we present X-SHIELD, a regularization of the T-SHIELD family which selects a specific transformation to enhance model explainability.
Finally, in Section~\ref{sec:technical_tshield}, we consider different technical aspects to be considered before applying any regularization of the T-SHIELD family.

\subsection{T-SHIELD regularization family}
\label{sec:tshield}

In this section, we define the T-SHIELD family~(Transformation-Selective Hidden Input Evaluation for Learning Dynamics), a family of regularizations designed to force a model to learn with less features. This is achieved by hiding part of the original input.

Formally, for $x\in X, x = (x_1,...,x_n)$, a certain hide transformation $T$ can be described as $T(X) = (T_1(x_1),...,T_n(x_n))$, where $T_i(x_i) = x_i$ if $x_i$ is one of the features not hided and  $T_i(x_i) = x_0$ if not, where $x_0$ is a neutral value that can be chosen. In Section~\ref{sec:technical_tshield}, we detail the complexities of choosing $x_0$ and our approach to choose it as a hiding technique.

For a certain transformation $T(x) = x'$, we can formally define a T-SHIELD regularization as:

$$
T-SHIELD(x,\Theta) = KL(f_\Theta(x'),f_\Theta(x)) + KL(f_\Theta(x),f_\Theta(x'))
$$

where $KL(-,-)$ is the Kullback-Leibler divergence between two random variables and, therefore, T-SHIELD is the simetricice KL divergence, also called Jeffreys divergence\citep{jeffreys1946invariant}, of both random variables.
The objective of a T-SHIELD regularization is to encourage the model to generalize without the whole example $x$, rewarding the model if it does not need too many features of $x$ to extract quality features.

A first approach of transformation $T$ can be described with a certain $\lambda \in [0,100]$ which controls the percentage of features to hide. With this $\lambda$ selected, the features can be randomly selected to be hidden. We call this approach R-SHIELD~(Random-SHIELD), and select indiscriminately $\lambda \%$ features to hide.

The definition of T-SHIELD regularization family has no theoretical restrictions except that the model must be differentiable. Besides, T-SHIELD regularization can be applied to any type of input, output, model, and task. 
It should also be noted that the use of T-SHIELD regularization can be classified as a Loss Augmentation method on the classification of Loss augmentation methods proposed in~\citep{weber2023beyond}. Also, unlike Saliency Guided~\citep{ismail2021improving}, T-SHIELD use a commutative loss function between the input and the modified input. This property makes our proposal to balance the influence of the input and the modified input instead of prioritizing that one of the outputs resembles the other.

\subsection{X-SHIELD: selection of the transformation T to enhance model explainability}
\label{sec:xshield_transformation}

In this section, we define the X-SHIELD regularization with the selection of a certain XAI-inspired transformation. 
The main goal of the proposed X-SHIELD transformation is to introduce knowledge by the action of hiding specific features. 
We use XAI techniques to hide the most appropriate features for this purpose, categorizing this proposal as a technique for improving the model with Loss Augmentation without human explanation labeling.

\begin{figure}[http]
    \centering
    \includegraphics[width=0.8\textwidth]{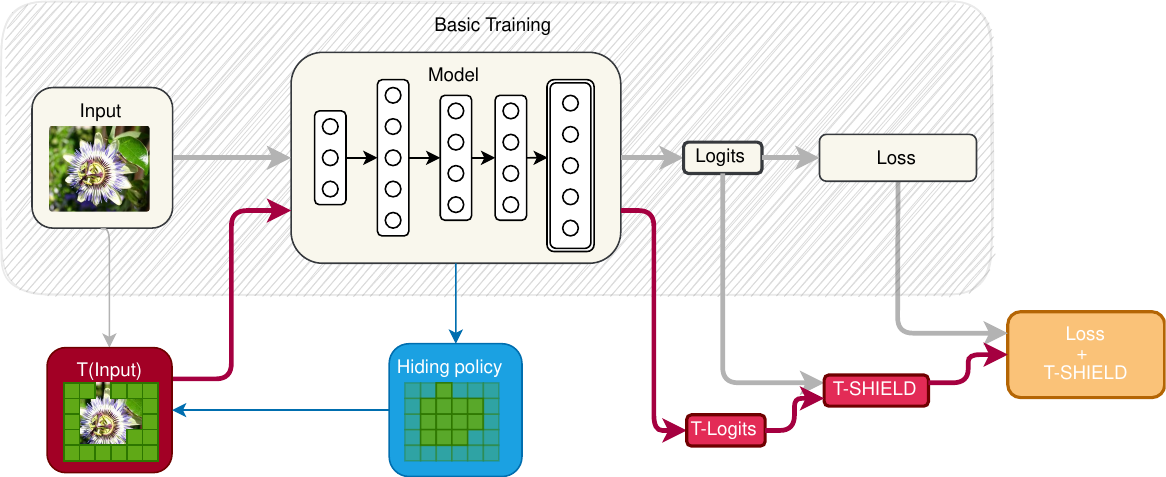}
    \caption{T-SHIELD workflow.}
    \label{fig:tshield_workflow}
\end{figure}

In Figure~\ref{fig:tshield_workflow}, we show the T-SHIELD workflow. We distinguish from the basic approach and the additional steps we add. 
Those steps are ordered as follows: 1. Obtention of the Feature Saliency map~(blue), 2. Hiding features from the input data~(red), 3. Computation of the Jeffreys divergence of the original and the transformed logits~(light red). 4. Final computation of the $Loss + T-SHIELD$ as the final loss to optimize~(orange).

The choice of the features to be hided is based on the estimation of its importance.
Formally, and as described in~\citep{sevillano2023revel}, the absolute importance of a feature $i$ in feature importance explanations can be estimated as 
$ \lVert\dfrac{\delta f_\Theta}{\delta V_i} \rVert$
where $V_i$ is the vector of the 
$i-th$ component of the importance matrix, and operator 
$ \lVert \cdot \rVert $ 
is the module operator of a vector. 
This factor is then the value of the differentiation of the model output with respect to the feature $i$.
For the sake of lower training time and simplicity, the approach chosen to determine the importance of each feature is the aforementioned approach. This approach is also implemented in different artificial intelligence libraries since it is used to the gradient descendant optimization algorithm.

With the previous definition of absolute importance of each feature $i$, let $T_{XAI}$ be the transformation $x' = T_{XAI}(x;\lambda)$ fixed to the value $x' = (x_0,x_0,..., x_0,x_i, x_{i+1},..., x_n)$. The $\lambda$ parameter is fixed in $\lambda < \dfrac{i}{n}\cdot100\%$ and the features $x_i$ are sorted by increasing absolute importance, thus making the first $\lambda\%$ least-important features to be hidden. Our regularization function can be formalized as follows:
$$
X-SHIELD(x,\Theta) = KL(f_\Theta(T_{XAI}(x;\lambda)),f_\Theta(x)) + KL(f_\Theta(x),f_\Theta(T_{XAI}(x;\lambda)))
$$
Note that this approach uses a general explanation of the model decision. This generalization allows it to be generalized as a regularization function in any task.

\begin{figure}
    \centering
    \includegraphics[width=0.5\textwidth]{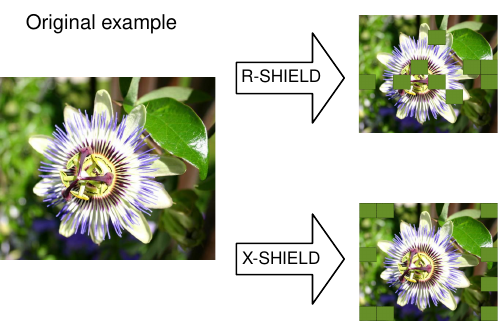}
    \caption{Transformations proposed by R-SHIELD and X-SHIELD respectively to an example of the Flowers dataset.}
    \label{fig:t_shield_transformations}
\end{figure}

In Figure~\ref{fig:t_shield_transformations}, we show an example of the Flowers dataset with the transformations of the previously defined in~\ref{sec:tshield} R-SHIELD transformation and X-SHIELD.
On the one hand, we appreciate that R-SHIELD chooses any feature indiscriminately from the example. All the features has the same probability on R-SHIELD to be hidden. 
On the other hand, X-SHIELD propose an informed transformation where the important features are not hidden. 
Intuitively, this transformation hide the noise of the background, letting the model to focus on the important part of the input data. In this example, the model should propose as important the whole flower, so it will not hide any part of it.

\subsection{Technical aspects for T-SHIELD regularizations}
\label{sec:technical_tshield}

Although we have defined T-SHIELD regularization family generally, and consequently X-SHIELD and R-SHIELD, there are several technical steps we must consider for each data type and possible subsequent proposals. These are: (i) Is there a difference between input variables and features? and (ii) How to hide a feature to the model if they are not prepared for that?
We take these considerations and adapt them to the image data type and to the proposal.

\paragraph{\hspace{20pt}(i) Difference between input variables and features? From pixels to features}
\label{sec:features}
When working on explainability, the concept of the input variable is not used. 
Instead, the concept of feature is more extended. 
The reason for using this concept is that we work with minimal units of knowledge. In the case of images, although a pixel is a minimum unit of information for images, it does not mean much to a human being. Grouping pixels, or superpixels, is often used as an image feature, as in LIME or SHAP XAI methods~\citep{ribeiro2016should,lundberg_unified_2017}. 
It is also used in explainable metrics~\citep{miro2023assessing}.

Different strategies have been used for the consideration of superpixels. 
A first approach is to use square segments of equal size~\citep{zhu2019guideline}. 
This approach does not require any extra computation since the segments are the same squares on every image.
On the other hand, different unsupervised segmentation methods have also been used to generate these regions~\citep{schallner2019effect}. 
Although even in the simplest cases where Deep Learning is not being used, minimal computation is needed to separate distinct regions. 
However, they usually identify significant regions.
We adopt the square segments approach since we want to save computational time in training.

\paragraph{\hspace{20pt}(ii) Hiding technique}
\label{sec:hide}
Once the concept of features in images is defined, we propose a method of hiding these features from the model we are working with. 
Following the formulation of \ref{sec:tshield}, the hiding technique can be understood as the $x_0$ by which to replace particular parts of the image.
This task is non-trivial due to the current lack of models specifically designed for this purpose.
A commonly quick and simple way used for this task is to change the region of the selected superpixels to a neutral color of choice, either black, white or the mean of the image or superpixel~\citep{amparore2021trust}. 
We implement in the proposal the method where we change the superpixel region to the mean color of the whole image.

\section{Experimental setup}
\label{sec:experimental_setup}

In this section, we describe the experimental setup we use in this work to empirically test the X-SHIELD regularization.
In Section~\ref{sec:benchmark_datasets}, we present the image datasets used as benchmarks. 
In Section~\ref{sec:models_optim}, we outline the different configurations of the experiments that we propose, considering basic configurations for the training, such the epochs, specific optimizer, validation scheme and evaluation scheme.
Finally, in Section~\ref{sec:performance}, we explain the different approaches we use to evaluate the different aspects of each model, both performance and explainability point of view.


\subsection{Benchmark selection}
\label{sec:benchmark_datasets}
The datasets selected as benchmarks are shown in Table~\ref{tab:datasets} with a small analytical description.
This selection is due partly to the fact that certain of these datasets have been used as benchmarks also in~\citep{yan2022towards} and~\citep{sevillano2023revel} for explainability tasks on image datasets. We added Flowers and Oxford IIIT Pet to study datasets with fewer examples. We also added the large-scale dataset of Imagenet 1K to test the effectiveness of this regularization on datasets with this amount of data.
There is variety in terms of image size, dataset size, number of classes, and black and white vs. color images which allows us to draw general conclusions from the study.

\begin{table}[http]
    \centering
    \begin{tabular}{ccccccc}
        \toprule
        Dataset & Nº classes & Original image size & Training & Test & RGB & Citation  \\ \hline
        CIFAR10 & 10 & $32\cdot32$ & 50.000 & 10.000 &Yes & \citep{cifar10}\\
        CIFAR100 & 100 & $32\cdot32$ & 50.000 & 10.000 & Yes & \citep{cifar100} \\
        FashionMNIST & 10 & $28\cdot28$ & 60.000 & 10.000 & No & \citep{xiao_fashion-mnist_2017} \\
        EMNIST-balanced & 47 & $28\cdot28$ & 112.800 & 18.800  & No & \citep{cohen2017emnist}\\
        Flowers & 102 & $ ~700 \cdot 500$ & 2040 & 6149 & Yes & \citep{nilsback2008automated}\\
        Oxford IIIT Pet & 37 & $500 \cdot ~ 300 $ & 3680 & 3669 & Yes & \citep{parkhi2012oxford}\\
        Imagenet 1k & 1000 & $224 \cdot 224$ & 1.281.167 & 100.000 & Yes & \citep{deng2009imagenet}\\
    \end{tabular}
    \caption{Descriptive table of the benchmarks selected}
    \label{tab:datasets}
\end{table}

\subsection{Models and optimization techniques}
\label{sec:models_optim}

As base models, we use Efficientnet~\citep{tan2019efficientnet} and Efficientnet V2~\citep{tan2021efficientnetv2} architectures since they are the state-of-the-art image classification datasets in their larger versions (Efficientnet B7 or Efficientnet V2-Medium). 
However, since the aim of this work is not to optimize the explanations, we decided to use lighter and faster models in terms of training time.
We choose Efficientnet B2 and Efficientnet V2 Small architectures to check whether the study is not dependent on a particular model.
We use Adam optimization technique~\citep{kingma2014adam} with L2 regularization for a $batch\_size$ of 32. To perform the experiments, we use a Tesla A100 GPU with 8 cores.

Each dataset has Test and Training sets separated, determined a priori by the specific dataset.
Each Training set has been separated into a 90\% training set and a 10\% validation set.
The training set is used to update the weights of the model on the training step. 
The validation set is used to validate every epoch of the loss function and maintain the best-results model weights for testing. We run each experiment for 80 epochs. We do not use this configuration of epoch in the Imagenet 1k training because of the amount of data it has and the computational cost: one epoch on Imagenet 1k consumes around 1 day. In this case, we consider an epoch to be updating weights of 5000 batches to decrease the computation time. The validation epoch has not been modified.

We choose to divide every image into 64 squares of the same size and generate 1000 different neighbours to generate the LIME explanation explanations. We set $\sigma$, a LIME variable that controls the weights of examples in terms of vicinity, to $8$. 10 different explanations for each example are generated to calculate the robustness metric. The selection of this settings has been previously used in~\citep{sevillano2023revel} for metrics evaluation purposes.

To test the performance of X-SHIELD and R-SHIELD, we use the same features as in LIME: we divide in 64 squares of the same size each image.
To study the influence of $\lambda$ on the X-SHIELD and R-SHIELD regularizations, we set $\lambda = 12\%, 24\%, 32\%, 48\%, 64\%$ and $72\%$ for Section~\ref{sec:lambda_study_xshield}.
As a result of this section, we conclude that a lower $\lambda$ value ends up in better results. For this reason, we then choose $\lambda = 3\%, 6\%, 9\%, 12\%, 15\%, 18\%, 21\%$ and $24\%$ for the rest of the experiments. Due to the computational time discussed in this section, for the experimentation with Imagenet 1k we have only considered $\lambda =3\%$ for X-SHIELD and R-SHIELD. This specific case is discussed separately.

\subsection{Performance evaluation}
\label{sec:performance}

We study the performance of loss and accuracy metrics on the test set for each experiment. 
To test the performance from the XAI perspective, we compare the explanations generated by LIME with the REVEL framework~\citep{sevillano2023revel} previously described in Section~\ref{sec:xAI_Ev}. 
We use 300 examples from the test set and generate 10 different explanations to obtain the REVEL framework evaluation. 
Then, we show violin plots of REVEL metrics for each dataset and regularization technique to compare the explanations generated. Plots are shown rather than numerical results to avoid overwhelming the reader and to simplify the analysis.

To check if the differences between results are statistically significant, we use Bayesian Signed-Rank test~\citep{carrasco2017rnpbst}, a non-parametric test that also allows us to add a Region Out of Practical Equivalence (ROPE) to determine if the differences are significant with an allowed margin of error.

\section{Experimental results}
\label{sec:experimental_results}

In this section, we delve into a detailed analysis of the results obtained from the experiments, aiming to provide a comprehensive evaluation of the use of the X-SHIELD regularization term. 
We show the results of each dataset individually and summarize the general analysis as a conclusion for each section.
In Setion~\ref{sec:lambda_study_xshield}, we study the impact of the $\lambda$ parameter on both X-SHIELD and R-SHIELD regularizations.
In Section~\ref{sec:test_performance}, we analyze the performance of X-SHIELD regularization on accuracy and loss metrics from a dual perspective: an analysis of whether the introduction of the X-SHIELD regularization improves the model and an analysis of whether the choice of the transformation T is important.
Finally, in Section~\ref{sec:visual_revel}, we analize the X-SHIELD performance over the REVEL metrics to test the model enhancement in explainability of this regularization.

\subsection{Impact of Nº of features hided~($\lambda$ parameter)}
\label{sec:lambda_study_xshield}

In this section, we analyze the impact of the $\lambda$ parameter of both X-SHIELD and R-SHIELD regularizations on the training and validation stages so we can make a fair comparison between them on the following sections, since the $\lambda$ parameter affects both regularizations. 
This comparison is made with the Efficientnet B2 model over the CIFAR10 dataset. 

\begin{figure}
    \centering
    \input{images/lambda_study/lambda_study_xshield.tex}
    \caption{Evolution of loss, accuracy and X-SHIELD regularization technique in the training and validation sets for Efficientnet B2 over CIFAR10 dataset with different $\lambda$ values. The "Basic" label refers to training the model without the X-SHIELD regularization term to compare}.
    \label{fig:lambda_study_xshield}
\end{figure}

\begin{figure}
    \centering
    \input{images/lambda_study/lambda_study_shield.tex}
    \caption{Evolution of loss, accuracy and R-SHIELD regularization in the training and validation sets for Efficientnet B2 over CIFAR10 dataset with different $\lambda$ values. The "Basic" label refers to training the model without the R-SHIELD regularization term to compare}.
    \label{fig:lambda_study_shield}
\end{figure}

We show in Figures~\ref{fig:lambda_study_xshield} and \ref{fig:lambda_study_shield}, the evolution graphs of loss, accuracy and regularization in the training and validation sets of X-SHIELD and R-SHIELD regularizations. 
We observe in both experiments that the lower the $\lambda$ value, the better the model's evolution graph in terms of loss and accuracy. 
This is expected since the $\lambda$ parameter controls how many features are hidden. 
Therefore, the lower the $\lambda$ value, the lower the difference between the original and the modified example is. This behavior leads to an earlier convergence. 
However, we notice that both regularizations decrease the convergence speed. The regularization terms adds complexity with the extra term to the loss function which explains this difficulty to learn.

The regularization term is higher when using X-SHIELD instead of R-SHIELD, as we may note in Figures~\ref{fig:lambda_study_regularization_val_xshield} and \ref{fig:lambda_study_regularization_val_shield}, where both graphics shows similar behavior with different scales. 
X-SHIELD regularization maximum value is around $0.4$ and R-SHIELD maximum value is around $0.3$. 
We must highlight that X-SHIELD regularization term is deterministic while R-SHIELD regularization term is sthocastic. 
As a consecuence, R-SHIELD gives the same relevance stochastically to all features, while X-SHIELD do not give relevance to the features that the model interpret as not important.

Decreasing $\lambda$ values improves the model's performance in both regularizations. Therefore, for the sake of fair comparison, the following experiments use values of $\lambda =  3, 6, 9, 12, 15, 18, 21, 24$ to compare X-SHIELD and R-SHIELD regularizations behavior.

\subsection{Performance analysis of X-SHIELD}
\label{sec:test_performance}
In this section, we analyze the performance of X-SHIELD regularization. 
The analysis is made in terms of accuracy and loss metrics on the test set for each dataset. 
To check if two experiments have significant differences, we use Bayesian Signed-Rank tests. 

The analysis has two well differentiated parts, depending on the aspect that we study.
On the first analysis, on Section~\ref{sec:Baseline_shield}, we test whether the introduction of the X-SHIELD regularization term in training improves model performance. 
On the second, on Section~\ref{sec:shield_xshield}, we check whether the design choice of the transformation T of X-SHIELD provides benefits with an enhancement in explainability.

\subsubsection{Introduction of the X-SHIELD regularization analysis}
\label{sec:Baseline_shield}

This section shows the accuracy and loss in test for each  dataset when using X-SHIELD with different $\lambda$'s. As in Section~\ref{sec:lambda_study_xshield}, we use $\lambda = 3, 6, 9, 12, 15, 18, 21, 24$ since higher values of $\lambda$ tend to worsen the model's performances.

Table~\ref{tab:loss_acc_regs_Baseline} shows the comparison of a basic training and the use of X-SHIELD in terms of accuracy and loss metrics.
We observe that, in terms of accuracy, the best configuration is achieved by X-SHIELD in 13 out of 14 experiments, with basic training achieving the best score in 1 case.
However, in terms of the loss metric, the basic training does not achieve the best score in any case. 
This might indicate that using X-SHIELD regularization makes de model generalize better.

\begin{table}[!http]\centering
    \scriptsize
    \caption{Accuracy and loss of basic training and the best X-SHIELD regularization configuration for Efficientnet B2 and Efficientnet V2 Small over all datasets. *Tested only with this value due to cost and time restriction.}
    \label{tab:loss_acc_regs_Baseline}
    \begin{tabular}{lrrrrrr}\toprule
\cellcolor[HTML]{999999}\textbf{Dataset} &\cellcolor[HTML]{999999}\textbf{Efficientnet} &\cellcolor[HTML]{999999}\textbf{Regularization} &\cellcolor[HTML]{999999}\textbf{$\lambda$} &\cellcolor[HTML]{999999}\textbf{Accuracy} &\cellcolor[HTML]{999999}\textbf{Loss} \\\midrule
\cellcolor[HTML]{f3f3f3}\textbf{} &\cellcolor[HTML]{f3f3f3}\textbf{} &\cellcolor[HTML]{f3f3f3}\textbf{Basic Training} &\cellcolor[HTML]{f3f3f3}0 &\cellcolor[HTML]{f3f3f3}\textbf{97.64\%} &\cellcolor[HTML]{f3f3f3}\textbf{0.0033} \\
\cellcolor[HTML]{f3f3f3}\textbf{CIFAR10} &\cellcolor[HTML]{f3f3f3}\textbf{B2} &\cellcolor[HTML]{cccccc}\textbf{X-SHIELD} &\cellcolor[HTML]{cccccc}3 &\cellcolor[HTML]{cccccc}97.59\% &\cellcolor[HTML]{cccccc}\textbf{0.0031} \\
\cellcolor[HTML]{f3f3f3}\textbf{} &\cellcolor[HTML]{cccccc}\textbf{} &\cellcolor[HTML]{f3f3f3}\textbf{Basic training} &\cellcolor[HTML]{f3f3f3}0 &\cellcolor[HTML]{f3f3f3}97.36\% &\cellcolor[HTML]{f3f3f3}0.0035 \\
\cellcolor[HTML]{f3f3f3}\textbf{} &\cellcolor[HTML]{cccccc}\textbf{V2 S} &\cellcolor[HTML]{cccccc}\textbf{X-SHIELD} &\cellcolor[HTML]{cccccc}12 &\cellcolor[HTML]{cccccc}\textbf{98.15\%} &\cellcolor[HTML]{cccccc}\textbf{0.0026} \\
\cellcolor[HTML]{cccccc}\textbf{} &\cellcolor[HTML]{f3f3f3}\textbf{} &\cellcolor[HTML]{f3f3f3}\textbf{Basic training} &\cellcolor[HTML]{f3f3f3}0 &\cellcolor[HTML]{f3f3f3}85.68\% &\cellcolor[HTML]{f3f3f3}0.0158 \\
\cellcolor[HTML]{cccccc}\textbf{CIFAR100} &\cellcolor[HTML]{f3f3f3}\textbf{B2} &\cellcolor[HTML]{cccccc}\textbf{X-SHIELD} &\cellcolor[HTML]{cccccc}9 &\cellcolor[HTML]{cccccc}\textbf{86.11\%} &\cellcolor[HTML]{cccccc}\textbf{0.0155} \\
\cellcolor[HTML]{cccccc}\textbf{} &\cellcolor[HTML]{cccccc}\textbf{} &\cellcolor[HTML]{f3f3f3}\textbf{Basic training} &\cellcolor[HTML]{f3f3f3}0 &\cellcolor[HTML]{f3f3f3}86.78\% &\cellcolor[HTML]{f3f3f3}0.0160 \\
\cellcolor[HTML]{cccccc}\textbf{} &\cellcolor[HTML]{cccccc}\textbf{V2 S} &\cellcolor[HTML]{cccccc}\textbf{X-SHIELD} &\cellcolor[HTML]{cccccc}6 &\cellcolor[HTML]{cccccc}\textbf{88.18\%} &\cellcolor[HTML]{cccccc}\textbf{0.0141} \\
\cellcolor[HTML]{f3f3f3}\textbf{} &\cellcolor[HTML]{f3f3f3}\textbf{} &\cellcolor[HTML]{f3f3f3}\textbf{Basic training} &\cellcolor[HTML]{f3f3f3}0 &\cellcolor[HTML]{f3f3f3}90.62\% &\cellcolor[HTML]{f3f3f3}0.0084 \\
\cellcolor[HTML]{f3f3f3}\textbf{EMNIST} &\cellcolor[HTML]{f3f3f3}\textbf{B2} &\cellcolor[HTML]{cccccc}\textbf{X-SHIELD} &\cellcolor[HTML]{cccccc}3 &\cellcolor[HTML]{cccccc}\textbf{91.06\%} &\cellcolor[HTML]{cccccc}\textbf{0.0080} \\
\cellcolor[HTML]{f3f3f3}\textbf{} &\cellcolor[HTML]{cccccc}\textbf{} &\cellcolor[HTML]{f3f3f3}\textbf{Basic training} &\cellcolor[HTML]{f3f3f3}0 &\cellcolor[HTML]{f3f3f3}90.46\% &\cellcolor[HTML]{f3f3f3}0.0085 \\
\cellcolor[HTML]{f3f3f3}\textbf{} &\cellcolor[HTML]{cccccc}\textbf{V2 S} &\cellcolor[HTML]{cccccc}\textbf{X-SHIELD} &\cellcolor[HTML]{cccccc}3 &\cellcolor[HTML]{cccccc}\textbf{90.78\%} &\cellcolor[HTML]{cccccc}\textbf{0.0082} \\
\cellcolor[HTML]{cccccc}\textbf{} &\cellcolor[HTML]{f3f3f3}\textbf{} &\cellcolor[HTML]{f3f3f3}\textbf{Basic training} &\cellcolor[HTML]{f3f3f3}0 &\cellcolor[HTML]{f3f3f3}93.90\% &\cellcolor[HTML]{f3f3f3}0.0057 \\
\cellcolor[HTML]{cccccc}\textbf{Fashion MNIST} &\cellcolor[HTML]{f3f3f3}\textbf{B2} &\cellcolor[HTML]{cccccc}\textbf{X-SHIELD} &\cellcolor[HTML]{cccccc}3 &\cellcolor[HTML]{cccccc}\textbf{95.31\%} &\cellcolor[HTML]{cccccc}\textbf{0.0047} \\
\cellcolor[HTML]{cccccc}\textbf{} &\cellcolor[HTML]{cccccc}\textbf{} &\cellcolor[HTML]{f3f3f3}\textbf{Basic training} &\cellcolor[HTML]{f3f3f3}0 &\cellcolor[HTML]{f3f3f3}93.94\% &\cellcolor[HTML]{f3f3f3}\textbf{0.0058} \\
\cellcolor[HTML]{cccccc}\textbf{} &\cellcolor[HTML]{cccccc}\textbf{V2 S} &\cellcolor[HTML]{cccccc}\textbf{X-SHIELD} &\cellcolor[HTML]{cccccc}3 &\cellcolor[HTML]{cccccc}\textbf{95.07\%} &\cellcolor[HTML]{cccccc}\textbf{0.0047} \\
\cellcolor[HTML]{f3f3f3}\textbf{} &\cellcolor[HTML]{f3f3f3}\textbf{} &\cellcolor[HTML]{f3f3f3}\textbf{Basic training} &\cellcolor[HTML]{f3f3f3}0 &\cellcolor[HTML]{f3f3f3}90.76\% &\cellcolor[HTML]{f3f3f3}\textbf{0.0120} \\
\cellcolor[HTML]{f3f3f3}\textbf{Flowers} &\cellcolor[HTML]{f3f3f3}\textbf{B2} &\cellcolor[HTML]{cccccc}\textbf{X-SHIELD} &\cellcolor[HTML]{cccccc}15 &\cellcolor[HTML]{cccccc}\textbf{92.47\%} &\cellcolor[HTML]{cccccc}\textbf{0.0088} \\
\cellcolor[HTML]{f3f3f3}\textbf{} &\cellcolor[HTML]{cccccc}\textbf{} &\cellcolor[HTML]{f3f3f3}\textbf{Basic training} &\cellcolor[HTML]{f3f3f3}0 &\cellcolor[HTML]{f3f3f3}92.44\% &\cellcolor[HTML]{f3f3f3}0.0091 \\
\cellcolor[HTML]{f3f3f3}\textbf{} &\cellcolor[HTML]{cccccc}\textbf{V2 S} &\cellcolor[HTML]{cccccc}\textbf{X-SHIELD} &\cellcolor[HTML]{cccccc}9 &\cellcolor[HTML]{cccccc}\textbf{94.21\%} &\cellcolor[HTML]{cccccc}\textbf{0.0069} \\
\cellcolor[HTML]{cccccc}\textbf{} &\cellcolor[HTML]{f3f3f3}\textbf{} &\cellcolor[HTML]{f3f3f3}\textbf{Basic training} &\cellcolor[HTML]{f3f3f3}0 &\cellcolor[HTML]{f3f3f3}90.32\% &\cellcolor[HTML]{f3f3f3}0.0115 \\
\cellcolor[HTML]{cccccc}\textbf{OxfordIIIT-Pet} &\cellcolor[HTML]{f3f3f3}\textbf{B2} &\cellcolor[HTML]{cccccc}\textbf{X-SHIELD} &\cellcolor[HTML]{cccccc}27 &\cellcolor[HTML]{cccccc}\textbf{91.28\%} &\cellcolor[HTML]{cccccc}\textbf{0.0106} \\
\cellcolor[HTML]{cccccc}\textbf{} &\cellcolor[HTML]{cccccc}\textbf{} &\cellcolor[HTML]{f3f3f3}\textbf{Basic training} &\cellcolor[HTML]{f3f3f3}0 &\cellcolor[HTML]{f3f3f3}92.59\% &\cellcolor[HTML]{f3f3f3}0.0111 \\
\cellcolor[HTML]{cccccc}\textbf{} &\cellcolor[HTML]{cccccc}\textbf{V2 S} &\cellcolor[HTML]{cccccc}\textbf{X-SHIELD} &\cellcolor[HTML]{cccccc}15 &\cellcolor[HTML]{cccccc}\textbf{93.35\%} &\cellcolor[HTML]{cccccc}\textbf{0.0075} \\

\cellcolor[HTML]{f3f3f3}\textbf{} &\cellcolor[HTML]{f3f3f3}\textbf{} &\cellcolor[HTML]{f3f3f3}\textbf{Basic training} &\cellcolor[HTML]{f3f3f3}0 &\cellcolor[HTML]{f3f3f3} 75.66\% &\cellcolor[HTML]{f3f3f3}\textbf{0.0314} \\
\cellcolor[HTML]{f3f3f3}\textbf{Imagenet 1K} &\cellcolor[HTML]{f3f3f3}\textbf{B2} &\cellcolor[HTML]{cccccc}\textbf{X-SHIELD} &\cellcolor[HTML]{cccccc}3* &\cellcolor[HTML]{cccccc}\textbf{76.09\%} &\cellcolor[HTML]{cccccc}\textbf{0.0295} \\
\cellcolor[HTML]{f3f3f3}\textbf{} &\cellcolor[HTML]{cccccc}\textbf{} &\cellcolor[HTML]{f3f3f3}\textbf{Basic training} &\cellcolor[HTML]{f3f3f3}0 &\cellcolor[HTML]{f3f3f3}78.87\% &\cellcolor[HTML]{f3f3f3}0.0279 \\
\cellcolor[HTML]{f3f3f3}\textbf{} &\cellcolor[HTML]{cccccc}\textbf{V2 S} &\cellcolor[HTML]{cccccc}\textbf{X-SHIELD} &\cellcolor[HTML]{cccccc}3* &\cellcolor[HTML]{cccccc}\textbf{79.89\%} &\cellcolor[HTML]{cccccc}\textbf{0.0249} \\
\bottomrule
\end{tabular}
    
\end{table}

\begin{figure}
    \centering
    \includegraphics[width=0.3\textwidth]{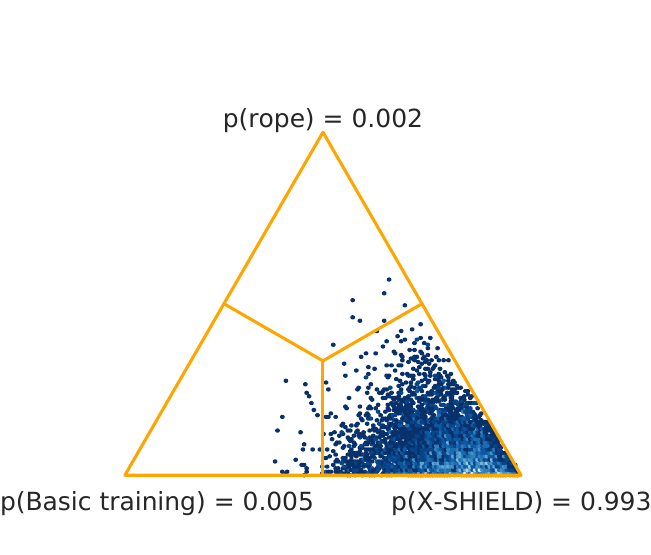}
    \caption{Bayesian Signed-Rank test of basic training vs X-SHIELD in terms of accuracy metric.}
    \label{fig:Bayes_test_Acc_Baseline_XSHIELD}
\end{figure}

In Figure~\ref{fig:Bayes_test_Acc_Baseline_XSHIELD}, we show the Bayesian Signed-Rank test which check wether the X-SHIELD improvement is significant. 
The p-value of the test is higher than 99\% so the differences betwen both experiments are relevant. 
This supports that X-SHIELD regularization enhance model quality.

\begin{figure}
    \centering
    \begin{subfigure}{0.4\textwidth}
        \includegraphics[width=\textwidth]{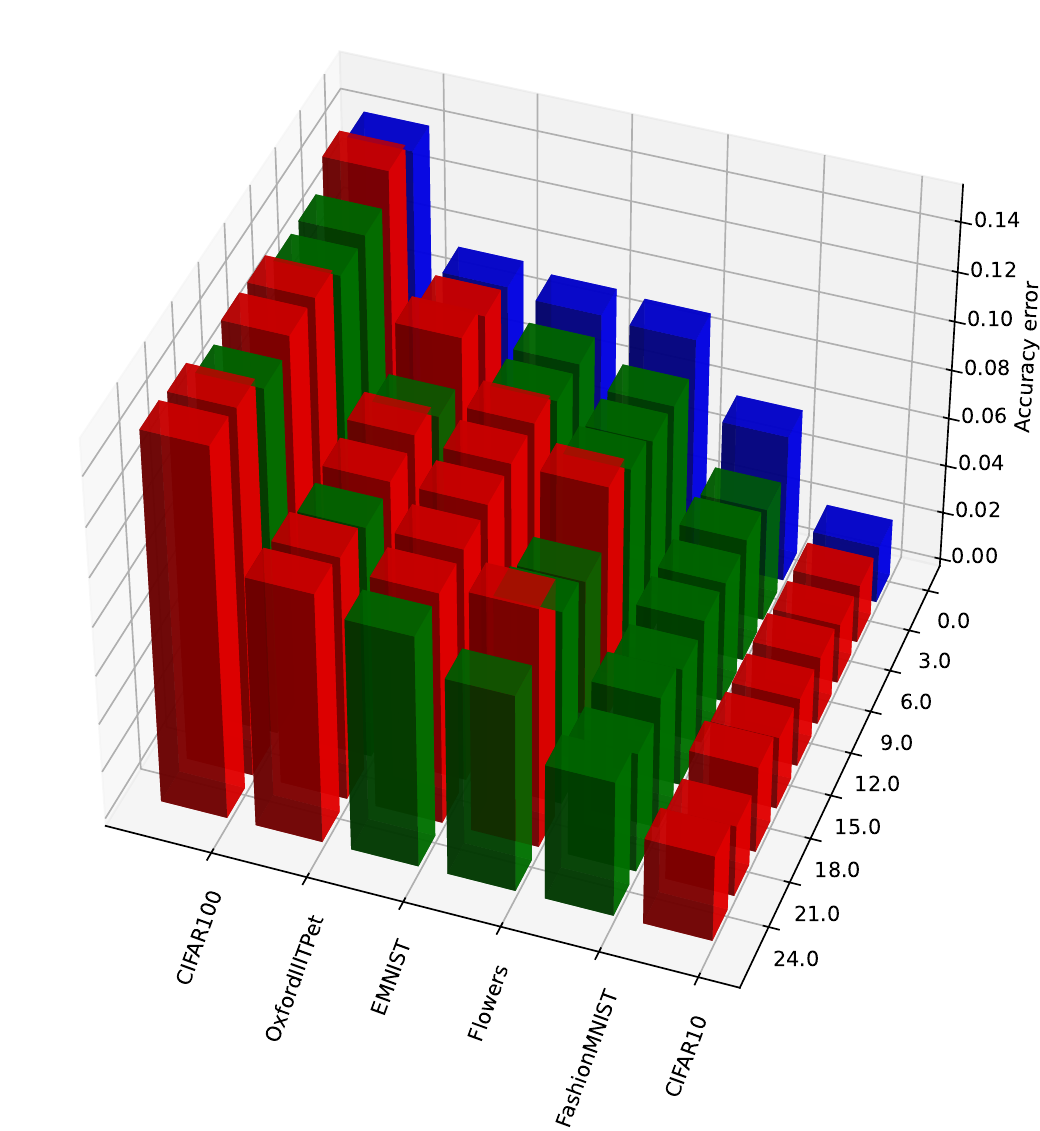}
        \caption{Efficientnet B2}
        \label{fig:3D_barplot_XSHIELD_B2}
    \end{subfigure}
    \begin{subfigure}{0.4\textwidth}
        \includegraphics[width=\textwidth]{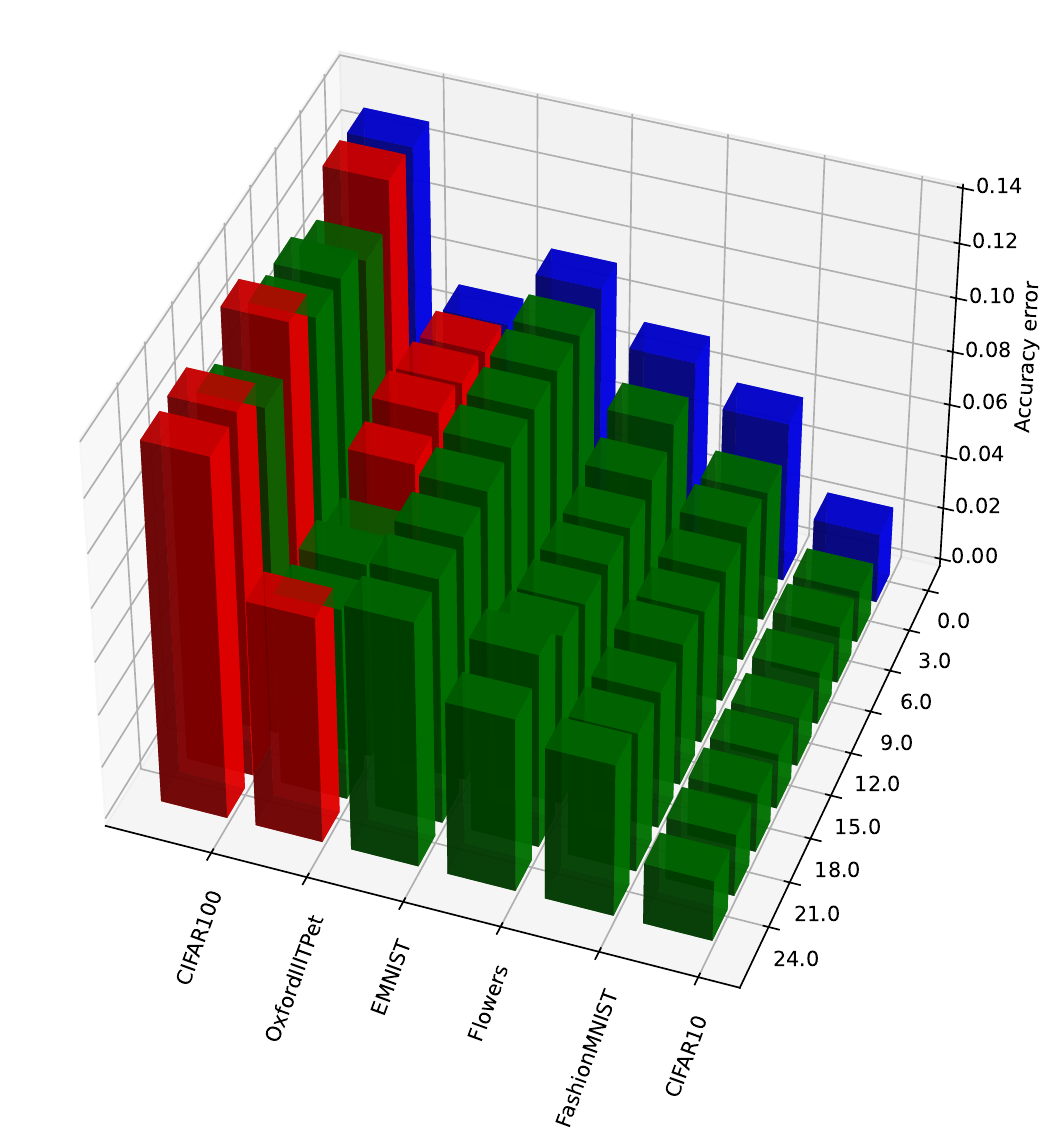}
        \caption{Efficientnet V2 Small}
        \label{fig:3D_barplot_XSHIELD_V2}
    \end{subfigure}
    \caption{3D barchart plot of the accuracy error rate of X-SHIELD regularization with different $\lambda$ values for Efficientnet B2 and V2 Small over the benchmark datasets. Blue color represents the basic training accuracy. Green and red colors represents if X-SHIELD accuracy with this value of $\lambda$ performs better or worse than Basic training, respectively. *Imagenet 1k is ocluded since experiments has been performed only with $\lambda=3$.}
    \label{fig:3D_barplot_XSHIELD}
\end{figure}

As a final visualization of the results, in Figure~\ref{fig:3D_barplot_XSHIELD} we show the 3D barchart plots with the $\lambda$'s values and the different accuracy error rates over different datasets but for Imagenet 1K due to cost and time restrictions. 

\begin{itemize}
    \item In Figure~\ref{fig:3D_barplot_XSHIELD_B2}, we notice that the use of X-SHIELD have different performance behavior. Depending on the dataset and the $\lambda$ parameter, X-SHIELD regularization may perform better or worse. Therefore, the selection of the $\lambda$ paarameter is essential to improve the model performance.
    \item In Figure~\ref{fig:3D_barplot_XSHIELD_V2}, we notice that X-SHIELD regularization tends to improve the model's performance in most cases. We note that in case of OxfordTIIIPet dataset is essential the selection of $\lambda$ parameter is to improve the model.
    \item Finally, and without leaving aside the Imagenet 1k dataset, we note in Table~\ref{tab:loss_acc_regs_Baseline} that X-SHIELD also performs better than the base training on this dataset with $\lambda=3$.
\end{itemize}

We observe that X-SHIELD has a better performance in Efficientnet V2 Small than Efficientnet B2. 

\subsubsection{Selection of the transformation T in performance enhancement}
\label{sec:shield_xshield}

In this section, we show the test results of accuracy and loss metrics in each of the datasets of X-SHIELD and R-SHIELD with different $\lambda$'s. 
This analysis is crucial to justify the use of the X-SHIELD transformation instead of other of the T-SHIELD family. R-SHIELD is a baseline to compare with because it removes the order of features proposed by the transformation of X-SHIELD.

As in previous Section~\ref{sec:Baseline_shield}, we use $\lambda = 3, 6, 9, 12, 15, 18, 21, 24$. 
As both regularizations have similar behavior with different $\lambda$ values, we use the best $\lambda$ value for each regularization to compare them.

In Table~\ref{tab:loss_acc_regs_x_shield}, we show the comparison of X-SHIELD and R-SHIELD in terms of accuracy and loss metrics. There isn't a clear difference between both regularizations, neither in terms of accuracy nor loss metrics. 
Examining the accuracy metric, we notice that X-SHIELD achieves the best results in 7 out of 12 cases, while R-SHIELD achieves the best score in 5.
Concerning the loss metric, X-SHIELD and R-SHIELD perform similarly, with 6 experiments yielding the best result each.

\begin{table}[!http]\centering
    \scriptsize
    \caption{Accuracy and loss of the best X-SHIELD and R-SHIELD regularization configuration for Efficientnet B2 and Efficientnet V2 Small over all datasets. *Tested only with this value due to cost and time restriction.}
    \label{tab:loss_acc_regs_x_shield}
    \begin{tabular}{lrrrrrr}\toprule
\cellcolor[HTML]{999999}\textbf{Dataset} &\cellcolor[HTML]{999999}\textbf{Model} &\cellcolor[HTML]{999999}\textbf{T-SHIELD} &\cellcolor[HTML]{999999}\textbf{$\lambda$} &\cellcolor[HTML]{999999}\textbf{Test/Accuracy} &\cellcolor[HTML]{999999}\textbf{Test/Loss} \\\midrule
\cellcolor[HTML]{f3f3f3}\textbf{} &\cellcolor[HTML]{f3f3f3}\textbf{} &\cellcolor[HTML]{f3f3f3}R &\cellcolor[HTML]{f3f3f3}3 &\cellcolor[HTML]{f3f3f3}\textbf{97.60\%} &\cellcolor[HTML]{f3f3f3}\textbf{0.0029} \\
\cellcolor[HTML]{f3f3f3}\textbf{CIFAR10} &\cellcolor[HTML]{f3f3f3}\textbf{B2} &\cellcolor[HTML]{cccccc}X &\cellcolor[HTML]{cccccc}3 &\cellcolor[HTML]{cccccc}97.59\% &\cellcolor[HTML]{cccccc}0.0031 \\
\cellcolor[HTML]{f3f3f3}\textbf{} &\cellcolor[HTML]{cccccc}\textbf{} &\cellcolor[HTML]{f3f3f3}R &\cellcolor[HTML]{f3f3f3}3 &\cellcolor[HTML]{f3f3f3}98.12\% &\cellcolor[HTML]{f3f3f3}0.0030 \\
\cellcolor[HTML]{f3f3f3}\textbf{} &\cellcolor[HTML]{cccccc}\textbf{V2 S} &\cellcolor[HTML]{cccccc}X &\cellcolor[HTML]{cccccc}12 &\cellcolor[HTML]{cccccc}\textbf{98.15\%} &\cellcolor[HTML]{cccccc}\textbf{0.0026} \\
\cellcolor[HTML]{cccccc}\textbf{} &\cellcolor[HTML]{f3f3f3}\textbf{} &\cellcolor[HTML]{f3f3f3}R &\cellcolor[HTML]{f3f3f3}3 &\cellcolor[HTML]{f3f3f3}\textbf{86.53\%} &\cellcolor[HTML]{f3f3f3}0.0157 \\
\cellcolor[HTML]{cccccc}\textbf{CIFAR100} &\cellcolor[HTML]{f3f3f3}\textbf{B2} &\cellcolor[HTML]{cccccc}X &\cellcolor[HTML]{cccccc}9 &\cellcolor[HTML]{cccccc}86.11\% &\cellcolor[HTML]{cccccc}\textbf{0.0155} \\
\cellcolor[HTML]{cccccc}\textbf{} &\cellcolor[HTML]{cccccc}\textbf{} &\cellcolor[HTML]{f3f3f3}R &\cellcolor[HTML]{f3f3f3}6 &\cellcolor[HTML]{f3f3f3}88.09\% &\cellcolor[HTML]{f3f3f3}0.0142 \\
\cellcolor[HTML]{cccccc}\textbf{} &\cellcolor[HTML]{cccccc}\textbf{V2 S} &\cellcolor[HTML]{cccccc}X &\cellcolor[HTML]{cccccc}6 &\cellcolor[HTML]{cccccc}\textbf{88.18\%} &\cellcolor[HTML]{cccccc}\textbf{0.0141} \\
\cellcolor[HTML]{f3f3f3}\textbf{} &\cellcolor[HTML]{f3f3f3}\textbf{} &\cellcolor[HTML]{f3f3f3}R &\cellcolor[HTML]{f3f3f3}15 &\cellcolor[HTML]{f3f3f3}90.94\% &\cellcolor[HTML]{f3f3f3}\textbf{0.0080} \\
\cellcolor[HTML]{f3f3f3}\textbf{EMNIST} &\cellcolor[HTML]{f3f3f3}\textbf{B2} &\cellcolor[HTML]{cccccc}X &\cellcolor[HTML]{cccccc}3 &\cellcolor[HTML]{cccccc}\textbf{91.06\%} &\cellcolor[HTML]{cccccc}\textbf{0.0080} \\
\cellcolor[HTML]{f3f3f3}\textbf{} &\cellcolor[HTML]{cccccc}\textbf{} &\cellcolor[HTML]{f3f3f3}R &\cellcolor[HTML]{f3f3f3}3 &\cellcolor[HTML]{f3f3f3}\textbf{90.91\%} &\cellcolor[HTML]{f3f3f3}\textbf{0.0079} \\
\cellcolor[HTML]{f3f3f3}\textbf{} &\cellcolor[HTML]{cccccc}\textbf{V2 S} &\cellcolor[HTML]{cccccc}X &\cellcolor[HTML]{cccccc}3 &\cellcolor[HTML]{cccccc}90.78\% &\cellcolor[HTML]{cccccc}0.0082 \\
\cellcolor[HTML]{cccccc}\textbf{} &\cellcolor[HTML]{f3f3f3}\textbf{} &\cellcolor[HTML]{f3f3f3}R &\cellcolor[HTML]{f3f3f3}18 &\cellcolor[HTML]{f3f3f3}95.18\% &\cellcolor[HTML]{f3f3f3}0.0050 \\
\cellcolor[HTML]{cccccc}\textbf{Fashion MNIST} &\cellcolor[HTML]{f3f3f3}\textbf{B2} &\cellcolor[HTML]{cccccc}X &\cellcolor[HTML]{cccccc}3 &\cellcolor[HTML]{cccccc}\textbf{95.31\%} &\cellcolor[HTML]{cccccc}\textbf{0.0047} \\
\cellcolor[HTML]{cccccc}\textbf{} &\cellcolor[HTML]{cccccc}\textbf{} &\cellcolor[HTML]{f3f3f3}R &\cellcolor[HTML]{f3f3f3}3 &\cellcolor[HTML]{f3f3f3}95.03\% &\cellcolor[HTML]{f3f3f3}\textbf{0.0047} \\
\cellcolor[HTML]{cccccc}\textbf{} &\cellcolor[HTML]{cccccc}\textbf{V2 S} &\cellcolor[HTML]{cccccc}X &\cellcolor[HTML]{cccccc}3 &\cellcolor[HTML]{cccccc}\textbf{95.07\%} &\cellcolor[HTML]{cccccc}\textbf{0.0047} \\
\cellcolor[HTML]{f3f3f3}\textbf{} &\cellcolor[HTML]{f3f3f3}\textbf{} &\cellcolor[HTML]{f3f3f3}R &\cellcolor[HTML]{f3f3f3}3 &\cellcolor[HTML]{f3f3f3}\textbf{92.54\%} &\cellcolor[HTML]{f3f3f3}\textbf{0.0088} \\
\cellcolor[HTML]{f3f3f3}\textbf{Flowers} &\cellcolor[HTML]{f3f3f3}\textbf{B2} &\cellcolor[HTML]{cccccc}X &\cellcolor[HTML]{cccccc}15 &\cellcolor[HTML]{cccccc}92.47\% &\cellcolor[HTML]{cccccc}\textbf{0.0088} \\
\cellcolor[HTML]{f3f3f3}\textbf{} &\cellcolor[HTML]{cccccc}\textbf{} &\cellcolor[HTML]{f3f3f3}R &\cellcolor[HTML]{f3f3f3}6 &\cellcolor[HTML]{f3f3f3}94.02\% &\cellcolor[HTML]{f3f3f3}0.0098 \\
\cellcolor[HTML]{f3f3f3}\textbf{} &\cellcolor[HTML]{cccccc}\textbf{V2 S} &\cellcolor[HTML]{cccccc}X &\cellcolor[HTML]{cccccc}9 &\cellcolor[HTML]{cccccc}\textbf{94.21\%} &\cellcolor[HTML]{cccccc}\textbf{0.0069} \\
\cellcolor[HTML]{cccccc}\textbf{} &\cellcolor[HTML]{f3f3f3}\textbf{} &\cellcolor[HTML]{f3f3f3}R &\cellcolor[HTML]{f3f3f3}12 &\cellcolor[HTML]{f3f3f3}90.81\% &\cellcolor[HTML]{f3f3f3}\textbf{0.0092} \\
\cellcolor[HTML]{cccccc}\textbf{OxfordIIIT-Pet} &\cellcolor[HTML]{f3f3f3}\textbf{B2} &\cellcolor[HTML]{cccccc}X &\cellcolor[HTML]{cccccc}27 &\cellcolor[HTML]{cccccc}\textbf{91.28\%} &\cellcolor[HTML]{cccccc}0.0106 \\
\cellcolor[HTML]{cccccc}\textbf{} &\cellcolor[HTML]{cccccc}\textbf{} &\cellcolor[HTML]{f3f3f3}R &\cellcolor[HTML]{f3f3f3}3 &\cellcolor[HTML]{f3f3f3}92.48\% &\cellcolor[HTML]{f3f3f3}0.0087 \\
\cellcolor[HTML]{cccccc}\textbf{} &\cellcolor[HTML]{cccccc}\textbf{V2 S} &\cellcolor[HTML]{cccccc}X &\cellcolor[HTML]{cccccc}15 &\cellcolor[HTML]{cccccc}\textbf{93.35\%} &\cellcolor[HTML]{cccccc}\textbf{0.0075} \\

\cellcolor[HTML]{f3f3f3}\textbf{} &\cellcolor[HTML]{f3f3f3}\textbf{} &\cellcolor[HTML]{f3f3f3}\textbf{R} &\cellcolor[HTML]{f3f3f3}0 &\cellcolor[HTML]{f3f3f3} 75.82\% &\cellcolor[HTML]{f3f3f3}\textbf{0.0298} \\
\cellcolor[HTML]{f3f3f3}\textbf{Imagenet 1K} &\cellcolor[HTML]{f3f3f3}\textbf{B2} &\cellcolor[HTML]{cccccc}\textbf{X} &\cellcolor[HTML]{cccccc}3* &\cellcolor[HTML]{cccccc}\textbf{75.86\%} &\cellcolor[HTML]{cccccc}\textbf{0.0295} \\
\cellcolor[HTML]{f3f3f3}\textbf{} &\cellcolor[HTML]{cccccc}\textbf{} &\cellcolor[HTML]{f3f3f3}\textbf{R} &\cellcolor[HTML]{f3f3f3}0 &\cellcolor[HTML]{f3f3f3}79.89\% &\cellcolor[HTML]{f3f3f3}0.0248 \\
\cellcolor[HTML]{f3f3f3}\textbf{} &\cellcolor[HTML]{cccccc}\textbf{V2 S} &\cellcolor[HTML]{cccccc}\textbf{X} &\cellcolor[HTML]{cccccc}3* &\cellcolor[HTML]{cccccc}\textbf{79.89\%} &\cellcolor[HTML]{cccccc}\textbf{0.0249} \\
\bottomrule
\end{tabular}
    
\end{table}

\begin{figure}
    \centering
    \includegraphics[width=0.4\textwidth]{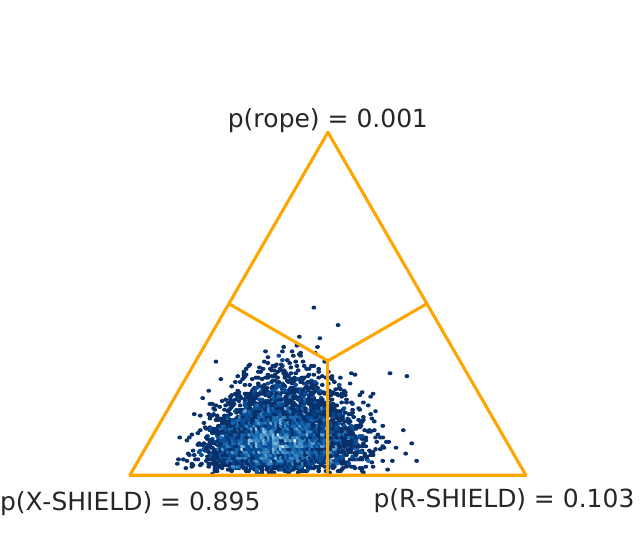}
    \caption{Bayesian Signed-Rank test of X-SHIELD vs R-SHIELD in terms of accuracy metric}
    \label{fig:Bayes_test_Acc_SHIELD_XSHIELD}
\end{figure}

To check if these experiments have significant differences, we show in Figure~\ref{fig:Bayes_test_Acc_SHIELD_XSHIELD} the Bayesian Signed-Rank tests. As both p-value are lower than 95\%, the differences between X-SHIELD and R-SHIELD are not significant, even if X-SHIELDS have a higher p-value of 0.895 against 0.103.
Then, we conclude that X-SHIELD and R-SHIELD has similar performance. 
However, we still need to evaluate the explainability perspective since it is the main improvement introduced by X-SHIELD regularization.

\subsection[Explainability analysis of X-SHIELD]{Explainability analysis of X-SHIELD: REVEL framework to quantitatively evaluate the model explainability enhancement}
\label{sec:visual_revel}

In this section, we analyze the impact of X-SHIELD regularization in terms of explainability. 
For this purpose, we use the REVEL framework metrics which provide a quantitative evaluation of the model's explanations. 
The metrics we analyze are described in Table~\ref{tab:metrics_summary}. 
For a detailed insights and analysis of each metric, we refer the reader to the original REVEL paper~\citep{sevillano2023revel}.

\begin{figure}[ht]
    \centering
    \begin{subfigure}{0.45\textwidth}
        \centering
        \includegraphics[width=\textwidth]{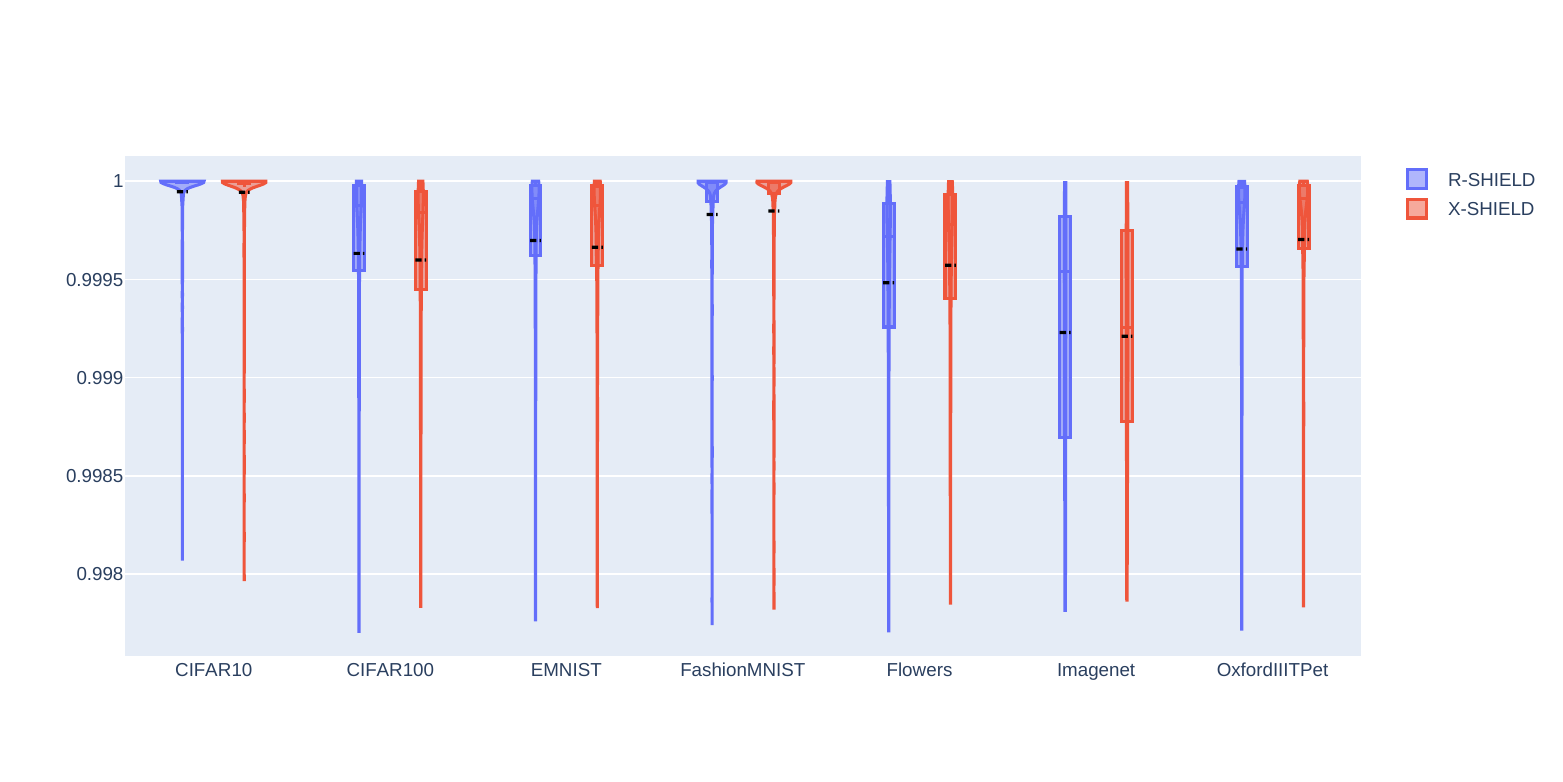}
        \caption{Local Concordance}
        \label{fig:Local_Concordance_violin}
    \end{subfigure}
    \begin{subfigure}{0.45\textwidth}
        \centering
        \includegraphics[width=\textwidth]{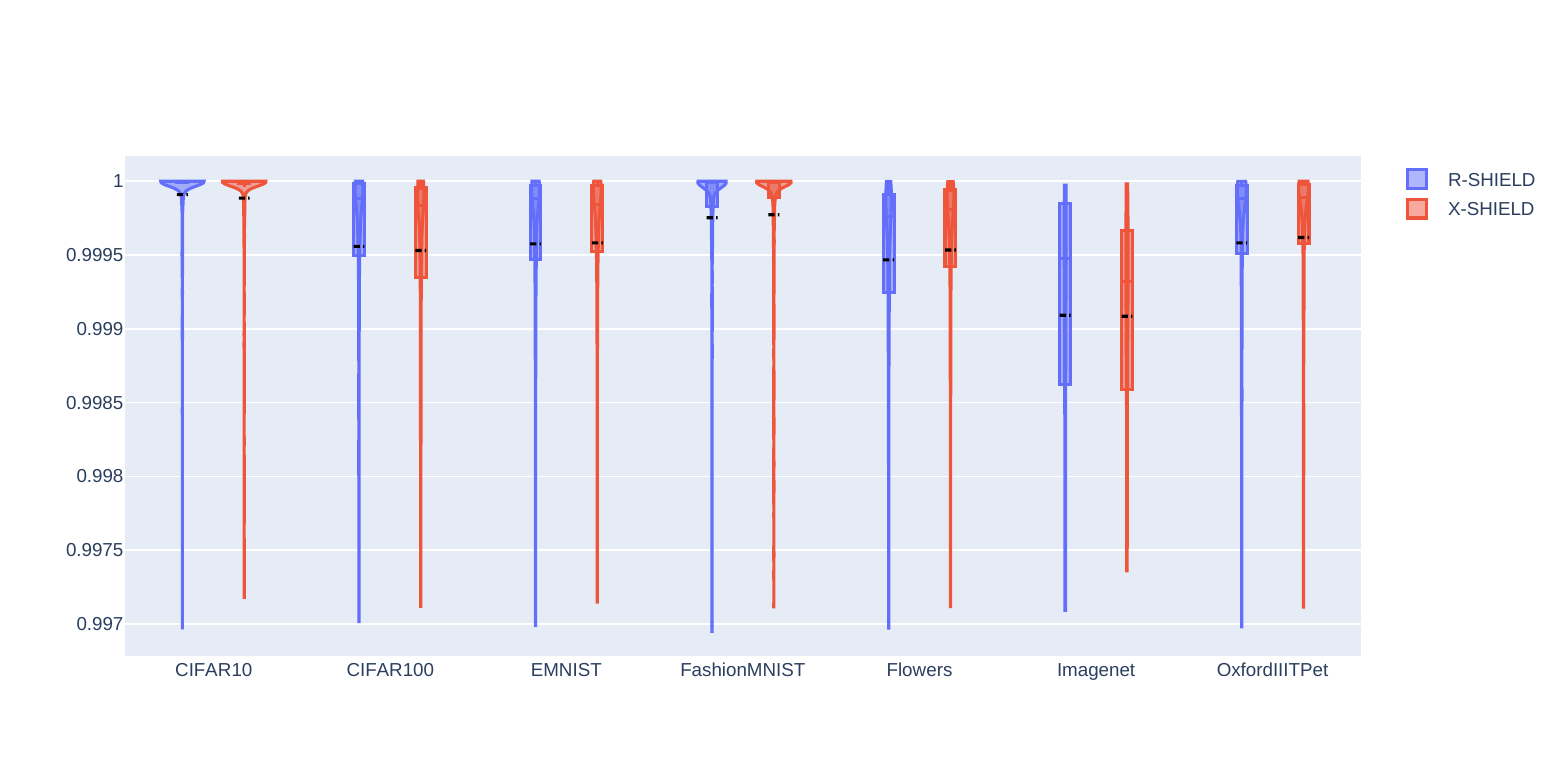}
        \caption{Local Fidelity}
        \label{fig:Local_Fidelity_violin}
    \end{subfigure}

    \begin{subfigure}{0.45\textwidth}
        \centering
        \includegraphics[width=\textwidth]{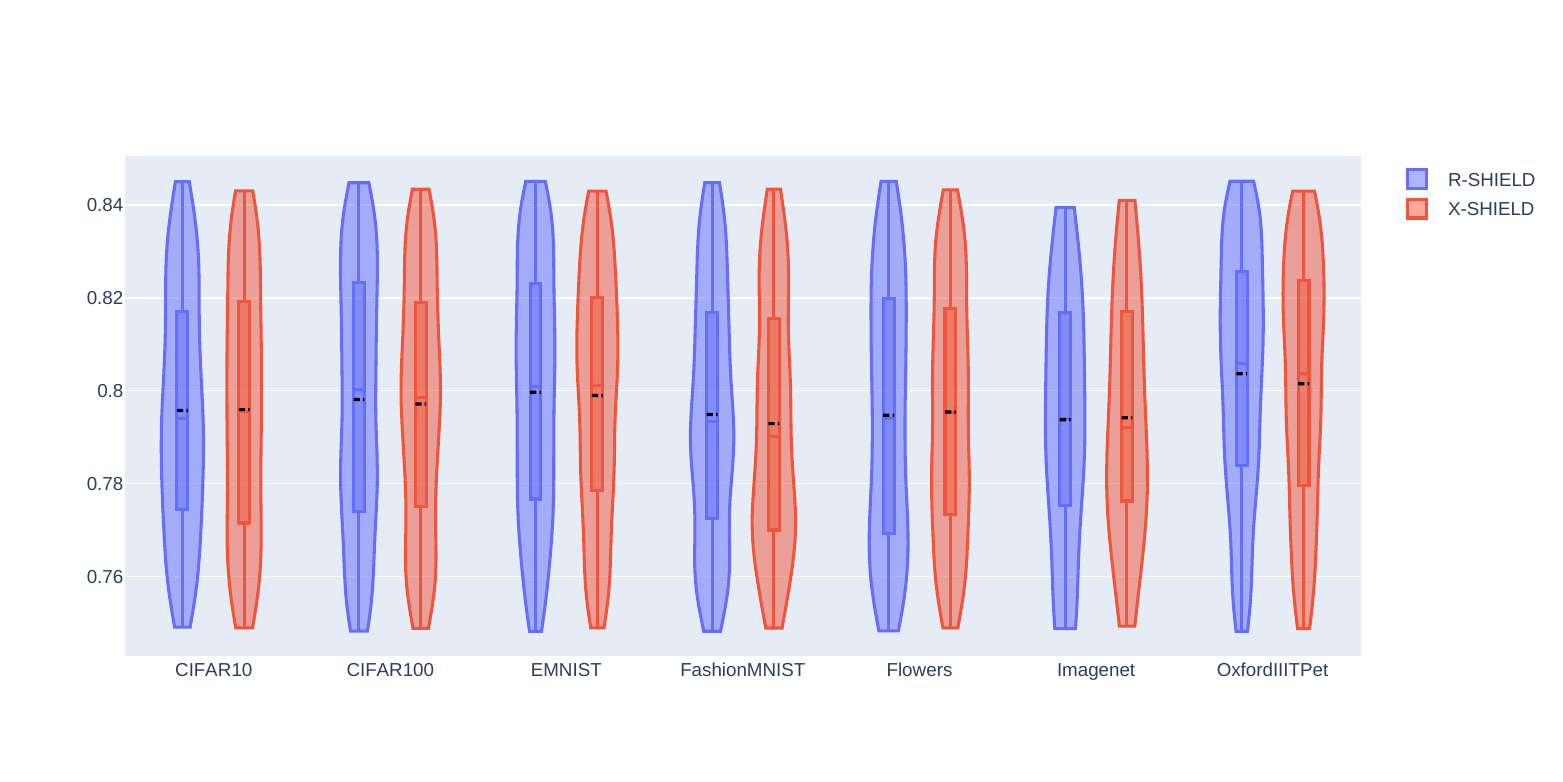}
        \caption{Conciseness}
        \label{fig:Conciseness_violin}
    \end{subfigure}
    \begin{subfigure}{0.45\textwidth}
        \centering
        \includegraphics[width=\textwidth]{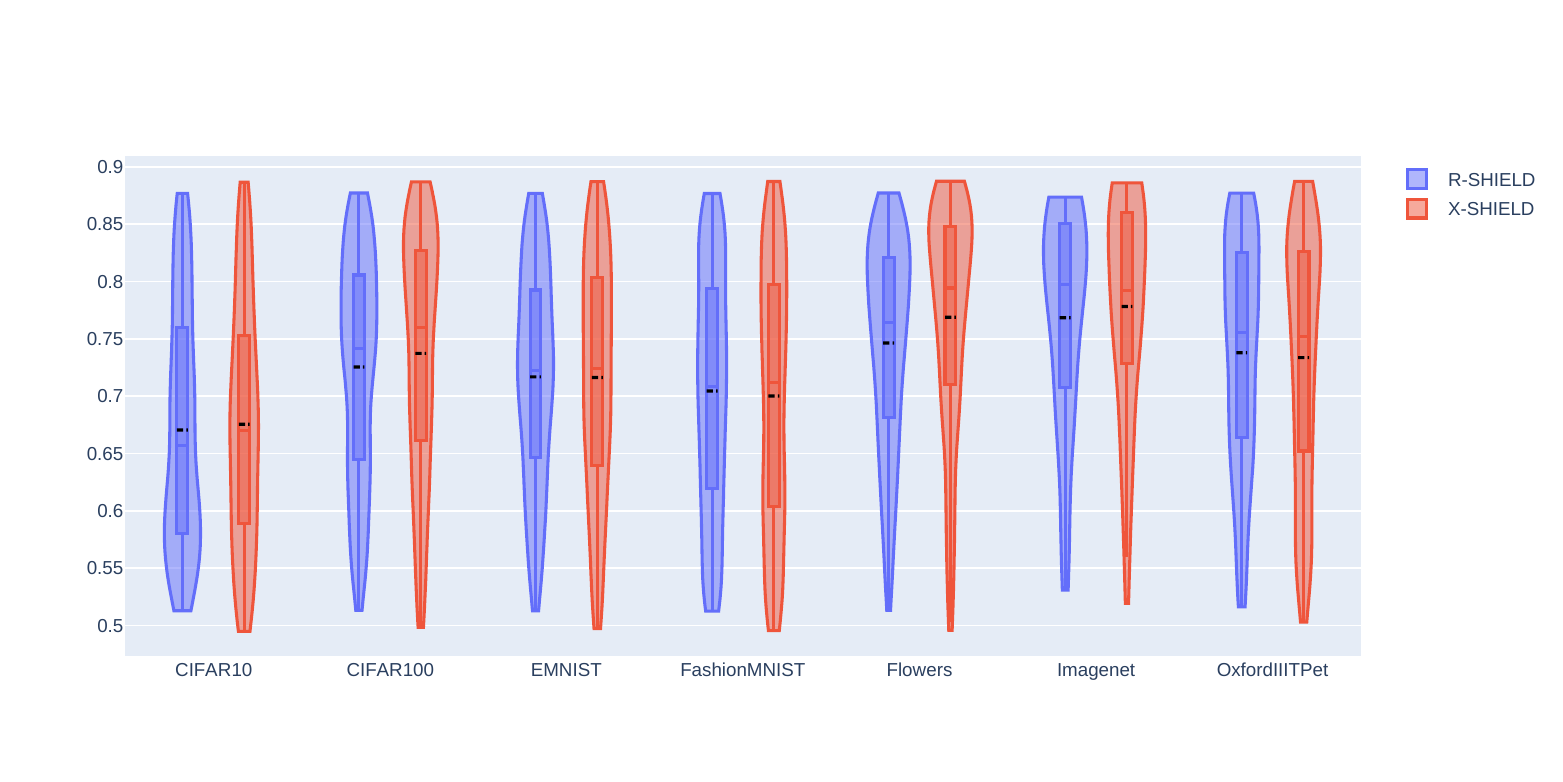}
        \caption{Prescriptivity}
        \label{fig:Prescriptivity_violin}
    \end{subfigure}

    \begin{subfigure}{0.45\textwidth}
        \centering
        \includegraphics[width=\textwidth]{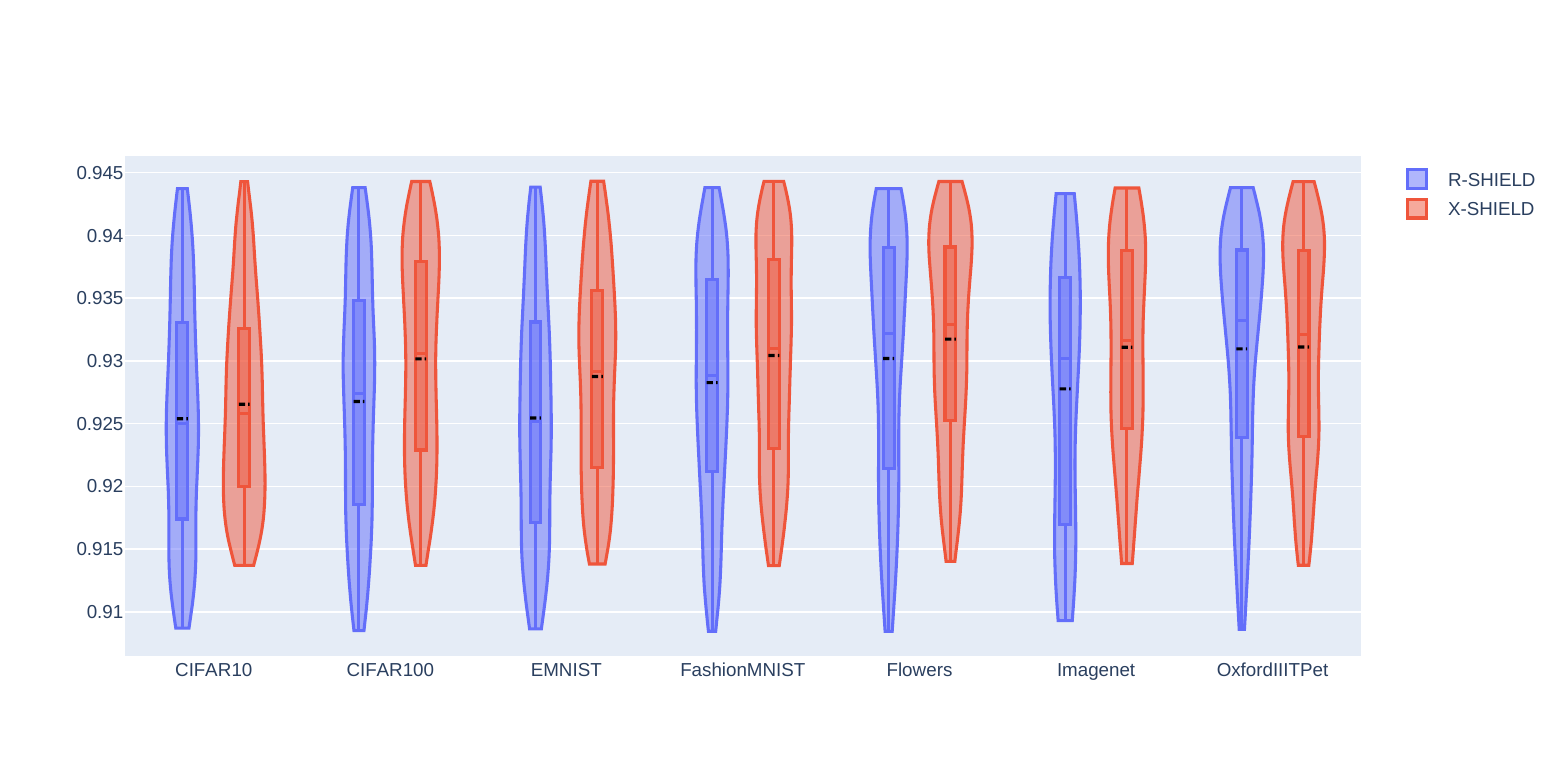}
        \caption{Robustness}
        \label{fig:Robustness_violin}
    \end{subfigure}
    \caption{Violin plots of the REVEL metrics comparing R-SHIELD and X-SHIELD regularizations.}
    \label{fig:Violin_plots}
\end{figure}

In Figure~\ref{fig:Violin_plots}, we show the distribution of each of the five metric over different datasets of X-SHIELD and R-SHIELD:

\begin{itemize}
    \item From Figures~\ref{fig:Local_Concordance_violin} and~\ref{fig:Local_Fidelity_violin}, we see that X-SHIELD and R-SHIELD regularization techniques shows similar results close to 1 on Local Concordance and Local Fidelity metrics, which is necessary to ensure the model's trustworthiness. 
    \item From Figure~\ref{fig:Conciseness_violin}, we observe that both models have similar behavior over the Conciseness metric. This means that both models have similar amount of features taken into  account to make predictions.
    \item From Figure~\ref{fig:Prescriptivity_violin}, we conclude that X-SHIELD regularization tends to have better Prescriptivity results than R-SHIELD regularization since the X-SHIELD median an 75\% quantil are higher than the R-SHIELD median an quantil in all datasets respectivaly. However, we may note that extreme values could be found in X-SHIELD distribution that worsen its performance.
    \item From Figure~\ref{fig:Robustness_violin}, we note that X-SHIELD also tends to have better Robustness metric. This is observed in the whole distribution in all datasets, where all the values represented on the violin graph~(max, min, mean and quantiles of 75\%,50\% and 25\%) of X-SHIELD are higher than their respective ones in R-SHIELD. This implies that X-SHIELD is more robust than R-SHIELD.
\end{itemize}

The violin plot visualization provides a general idea of the behavior of the models in terms of the REVEL metrics distribution. 
However, we refer to statistical tests to ensure that the observed differences are significant.
We provide Bayesian Signed-Rank test to evaluate whether there are relevant differences betwen X-SHIELD and R-SHIELD experiments on each metric. 

\begin{figure}[ht]
    \centering
    \begin{subfigure}{0.30\textwidth}
        \centering
        \includegraphics[width=\textwidth]{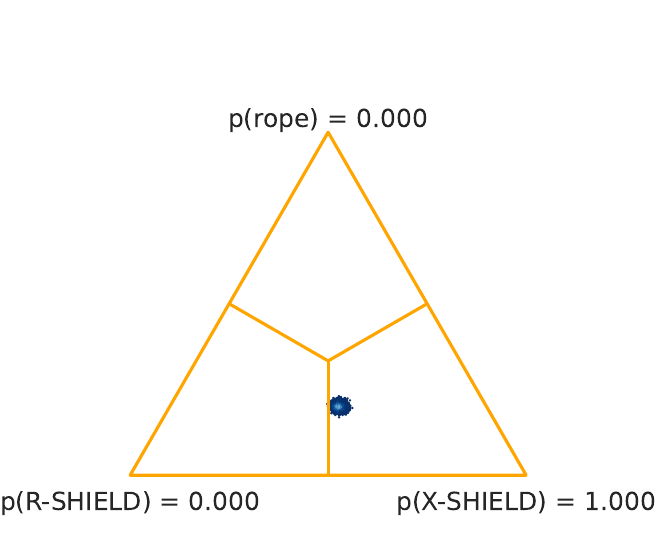}
        \caption{Local Concordance}
        \label{fig:Local_Concordance_best}
    \end{subfigure}
    \begin{subfigure}{0.30\textwidth}
        \centering
        \includegraphics[width=\textwidth]{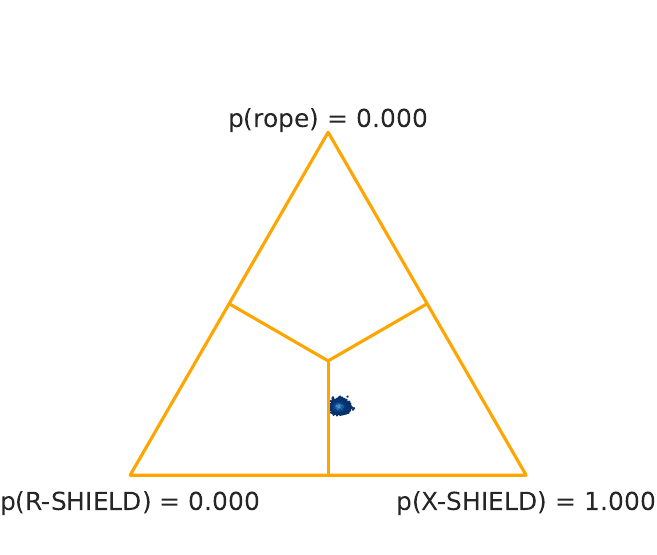}
        \caption{Local Fidelity}
        \label{fig:Local_Fidelity_best}
    \end{subfigure}
    \begin{subfigure}{0.30\textwidth}
        \centering
        \includegraphics[width=\textwidth]{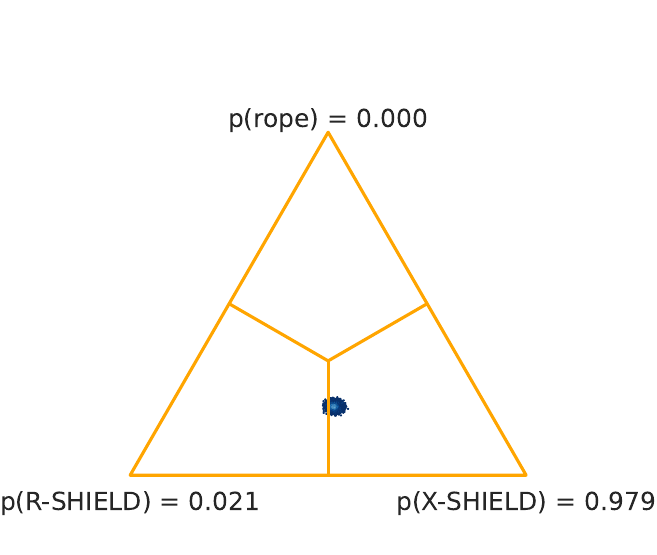}
        \caption{Conciseness}
        \label{fig:Conciseness_best}
    \end{subfigure}

    \begin{subfigure}{0.30\textwidth}
        \centering
        \includegraphics[width=\textwidth]{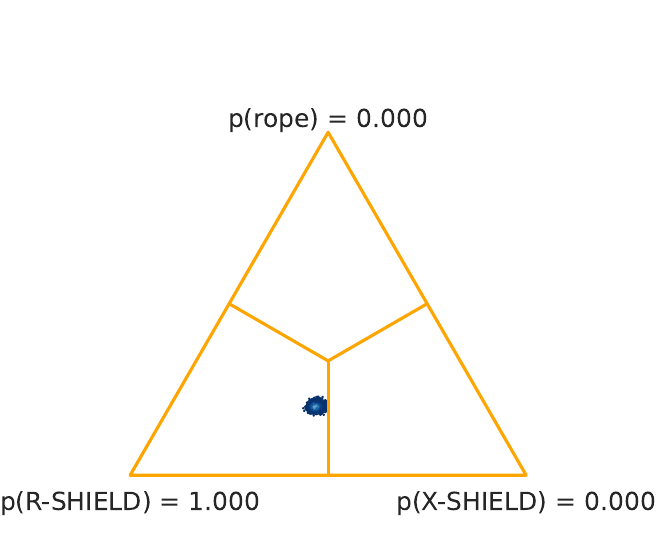}
        \caption{Prescriptivity}
        \label{fig:Prescriptivity_best}
    \end{subfigure}
    \begin{subfigure}{0.30\textwidth}
        \centering
        \includegraphics[width=\textwidth]{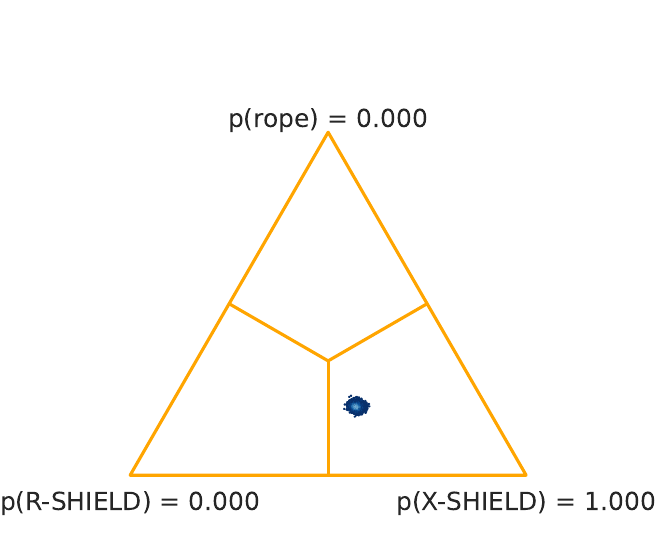}
        \caption{Robustness}
        \label{fig:Robustness_best}
    \end{subfigure}
    \caption{Bayesian Signed-Rank test for the REVEL metrics comparing R-SHIELD and X-SHIELD regularizations.}
    \label{fig:Bayes_revel}
\end{figure}

In Figure~\ref{fig:Bayes_revel},  we show the Bayesian Signed-Rank test for each metric comparing the behavior of using X-SHIELD or R-SHIELD regularization. We notice that all of them are not as disperse as the previous accuracy bayesian test. This lower dispersion is due to the number of examples used in the REVEL metrics, calculated for each example of the dataset. The accuracy metric is calculated for the whole test set so they have less examples. 
The lower dispersion with higher number of examples concordates with the work on~\citep{ghosal1997review}, which states that the posterior distribution tends to a fixed value as the number of examples increases.

We observe that there are significant differences between X-SHIELD and R-SHIELD, with X-SHIELD performing better in terms of Local Concordance, Local Fidelity, Conciseness and Robustness metrics, with a p-value of less than $0.02$ in the worst case. R-SHIELD, on the other hand, performs better in terms of Prescriptivity metric. 

We can summarize each metric result and their implications as follows: 

\begin{itemize}
    \item \textbf{Local Concordance and Local Fidelity:} X-SHIELD regularization tends to have better Local Concordance and Local Fidelity metrics than R-SHIELD regularization but these metrics are both close to the maximum value. Since the purpose of Local Concordance and Local Fidelity is to ensure that explanations of the model does not contradict the model's predictions, we do not emphasize the significant differences between X-SHIELD and R-SHIELD on these metrics.
    \item \textbf{Conciseness:} X-SHIELD regularization tends to discard the least relevant features. This is reflected in a higher Conciseness metric, since it means that the model needs to use fewer features to make a prediction.
    \item \textbf{Prescriptivity:} R-SHIELD regularization generates a neighbor of the input data randomly on each training example. X-SHIELD also generates a neighbor of the input data, but it is biased so the least relevant features are hidden. Thus, R-SHIELD tends to have better knowledge of the complete neighborhood, not just the non-relevant features modifications. This is reflected in a higher Prescriptivity metric.
    \item \textbf{Robustness:} X-SHIELD regularization generates a neighbor of the input data biased so the least relevant features are hidden. Then, the explanations generated by models trained with X-SHIELD gives less importance to the least relevant features, ensuring the model's explanations to be focused on the relevant ones ensuring a robust explanations, reflected in a higher Robustness metric. 
\end{itemize}

In conclusion, while R-SHIELD performs well in achieving an improvement in performance, X-SHIELD in addition improves most explainability aspects, being this regularization the one that is recommended.

\section{Conclusions}
\label{sec:conclusions}

This work introduces T-SHIELD, a regularization family  classified as a Loss Augmentation XAI-based method developed to improve model quality by hiding features.  Within this family and aligned with the Red XAI perspective, the study also introduce X-SHIELD, an informed XAI-based T-SHIELD member which selects wisely the features to hide. We analyze the X-SHIELD approximation with basic training and another uninformed approximation of the T-SHIELD family, R-SHIELD.

The results obtained support the initial hypothesis that the combination of the regularization perspective and the XAI perspective improves the explainability and quality of the model using X-SHIELD regularization. 
X-SHIELD regularization is an useful regularization tool that can be integrated into any type of training framework and decreases overfitting, enhance model performance and model explainability.

X-SHIELD regularization presents a significant contribution to addressing the duality between explainability and model quality, providing a valuable tool for researchers and practitioners in the field of AI. 

This proposal and study opens potential window to hiding image features for learning. As future work, different proposals of T-SHIELD regularization should be explored, such as the use of other methods of hiding features or by a new transformation T which improves the performance or the explainability aspect of the models.

\section*{Acknowledgments}

This work was supported by the Spanish Ministry of Science and Technology under project  	PID2023-150070NB-I00 financed by MCIN/AEI/10.13039/501100011033. 

\bibliographystyle{unsrtnat}
\bibliography{references} 

\begin{thebibliography}{48}
\providecommand{\natexlab}[1]{#1}
\providecommand{\url}[1]{\texttt{#1}}
\expandafter\ifx\csname urlstyle\endcsname\relax
  \providecommand{\doi}[1]{doi: #1}\else
  \providecommand{\doi}{doi: \begingroup \urlstyle{rm}\Url}\fi

\bibitem[Arrieta et~al.(2020)Arrieta, D{\'\i}az-Rodr{\'\i}guez, Del~Ser, Bennetot, Tabik, Barbado, Garc{\'\i}a, Gil-L{\'o}pez, Molina, Benjamins, et~al.]{arrieta2020explainable}
Alejandro~Barredo Arrieta, Natalia D{\'\i}az-Rodr{\'\i}guez, Javier Del~Ser, Adrien Bennetot, Siham Tabik, Alberto Barbado, Salvador Garc{\'\i}a, Sergio Gil-L{\'o}pez, Daniel Molina, Richard Benjamins, et~al.
\newblock Explainable artificial intelligence ({XAI}): Concepts, taxonomies, opportunities and challenges toward responsible {AI}.
\newblock \emph{Information Fusion}, 58:\penalty0 82--115, 2020.

\bibitem[Wachter et~al.(2017)Wachter, Mittelstadt, and Russell]{wachter2017counterfactual}
Sandra Wachter, Brent Mittelstadt, and Chris Russell.
\newblock Counterfactual explanations without opening the black box: Automated decisions and the {GDPR}.
\newblock \emph{Harvard Journal of Law \& Technology}, 31:\penalty0 841--888, 2017.

\bibitem[Ribeiro et~al.(2016)Ribeiro, Singh, and Guestrin]{ribeiro2016should}
Marco~Tulio Ribeiro, Sameer Singh, and Carlos Guestrin.
\newblock "{Why Should I Trust You?}": Explaining the predictions of any classifier.
\newblock In \emph{Proceedings of the 22nd ACM SIGKDD International Conference on Knowledge Discovery and Data Mining}, KDD '16, page 1135–1144, New York, NY, USA, 2016. Association for Computing Machinery.

\bibitem[Lundberg and Lee(2017)]{lundberg_unified_2017}
Scott~M. Lundberg and Su-In Lee.
\newblock A unified approach to interpreting model predictions.
\newblock In \emph{Proceedings of the 31st international conference on neural information processing systems}, pages 4768--4777, 2017.

\bibitem[Shrikumar et~al.(2017)Shrikumar, Greenside, and Kundaje]{shrikumar2017learning}
Avanti Shrikumar, Peyton Greenside, and Anshul Kundaje.
\newblock Learning important features through propagating activation differences.
\newblock In \emph{International conference on machine learning}, pages 3145--3153. Proceedings of Machine Learning Research, 2017.

\bibitem[Sundararajan et~al.(2017)Sundararajan, Taly, and Yan]{sundararajan2017axiomatic}
Mukund Sundararajan, Ankur Taly, and Qiqi Yan.
\newblock Axiomatic attribution for deep networks.
\newblock In \emph{International conference on machine learning}, pages 3319--3328, 2017.

\bibitem[Bennetot et~al.(2024)Bennetot, Donadello, El~Qadi El~Haouari, Dragoni, Frossard, Wagner, Sarranti, Tulli, Trocan, Chatila, et~al.]{bennetot2024practical}
Adrien Bennetot, Ivan Donadello, Ayoub El~Qadi El~Haouari, Mauro Dragoni, Thomas Frossard, Benedikt Wagner, Anna Sarranti, Silvia Tulli, Maria Trocan, Raja Chatila, et~al.
\newblock A practical tutorial on explainable ai techniques.
\newblock \emph{ACM Computing Surveys}, 2024.

\bibitem[Ali et~al.(2023)Ali, Abuhmed, El-Sappagh, Muhammad, Alonso-Moral, Confalonieri, Guidotti, {Del Ser}, Díaz-Rodríguez, and Herrera]{ALI2023101805}
Sajid Ali, Tamer Abuhmed, Shaker El-Sappagh, Khan Muhammad, Jose~M. Alonso-Moral, Roberto Confalonieri, Riccardo Guidotti, Javier {Del Ser}, Natalia Díaz-Rodríguez, and Francisco Herrera.
\newblock Explainable artificial intelligence (xai): What we know and what is left to attain trustworthy artificial intelligence.
\newblock \emph{Information Fusion}, 99:\penalty0 101805, 2023.
\newblock ISSN 1566-2535.
\newblock \doi{https://doi.org/10.1016/j.inffus.2023.101805}.
\newblock URL \url{https://www.sciencedirect.com/science/article/pii/S1566253523001148}.

\bibitem[Kong et~al.(2024)Kong, Liu, and Zhu]{kong2024toward}
Xiangwei Kong, Shujie Liu, and Luhao Zhu.
\newblock Toward human-centered xai in practice: A survey.
\newblock \emph{Machine Intelligence Research}, pages 1--31, 2024.

\bibitem[Longo et~al.(2024)Longo, Brcic, Cabitza, Choi, Confalonieri, Del~Ser, Guidotti, Hayashi, Herrera, Holzinger, et~al.]{longo2024explainable}
Luca Longo, Mario Brcic, Federico Cabitza, Jaesik Choi, Roberto Confalonieri, Javier Del~Ser, Riccardo Guidotti, Yoichi Hayashi, Francisco Herrera, Andreas Holzinger, et~al.
\newblock Explainable artificial intelligence ({XAI}) 2.0: A manifesto of open challenges and interdisciplinary research directions.
\newblock \emph{Information Fusion}, page 102301, 2024.

\bibitem[Biecek and Samek(2024)]{biecekposition}
Przemyslaw Biecek and Wojciech Samek.
\newblock Position: Explain to question not to justify.
\newblock In \emph{Under review at the Forty-First International Conference on Machine Learning}, 2024.

\bibitem[Amparore et~al.(2021)Amparore, Perotti, and Bajardi]{amparore2021trust}
Elvio Amparore, Alan Perotti, and Paolo Bajardi.
\newblock To trust or not to trust an explanation: using {LEAF} to evaluate local linear {XAI} methods.
\newblock \emph{PeerJ Computer Science}, 7:\penalty0 1--26, 2021.

\bibitem[Sevillano-García et~al.(2023)Sevillano-García, Luengo, and Herrera]{sevillano2023revel}
Iván Sevillano-García, Julián Luengo, and Francisco Herrera.
\newblock {REVEL} framework to measure local linear explanations for black-box models: Deep learning image classification case study.
\newblock \emph{International Journal of Intelligent Systems}, pages 1--34, 2023.

\bibitem[Moradi et~al.(2020)Moradi, Berangi, and Minaei]{moradi2020survey}
Reza Moradi, Reza Berangi, and Behrouz Minaei.
\newblock A survey of regularization strategies for deep models.
\newblock \emph{Artificial Intelligence Review}, 53:\penalty0 3947--3986, 2020.

\bibitem[Weber et~al.(2023)Weber, Lapuschkin, Binder, and Samek]{weber2023beyond}
Leander Weber, Sebastian Lapuschkin, Alexander Binder, and Wojciech Samek.
\newblock Beyond explaining: Opportunities and challenges of xai-based model improvement.
\newblock \emph{Information Fusion}, 92:\penalty0 154--176, 2023.

\bibitem[Angelov et~al.(2021)Angelov, Soares, Jiang, Arnold, and Atkinson]{angelov2021explainable}
Plamen~P Angelov, Eduardo~A Soares, Richard Jiang, Nicholas~I Arnold, and Peter~M Atkinson.
\newblock Explainable artificial intelligence: an analytical review.
\newblock \emph{Wiley Interdisciplinary Reviews: Data Mining and Knowledge Discovery}, 11\penalty0 (5):\penalty0 e1424, 2021.

\bibitem[Bodria et~al.(2023)Bodria, Giannotti, Guidotti, Naretto, Pedreschi, and Rinzivillo]{bodria2023benchmarking}
Francesco Bodria, Fosca Giannotti, Riccardo Guidotti, Francesca Naretto, Dino Pedreschi, and Salvatore Rinzivillo.
\newblock Benchmarking and survey of explanation methods for black box models.
\newblock \emph{Data Mining and Knowledge Discovery}, 37\penalty0 (5):\penalty0 1719--1778, 2023.

\bibitem[Selvaraju et~al.(2017)Selvaraju, Cogswell, Das, Vedantam, Parikh, and Batra]{selvaraju2017grad}
Ramprasaath~R Selvaraju, Michael Cogswell, Abhishek Das, Ramakrishna Vedantam, Devi Parikh, and Dhruv Batra.
\newblock Grad-cam: Visual explanations from deep networks via gradient-based localization.
\newblock In \emph{Proceedings of the IEEE international conference on computer vision}, pages 618--626, 2017.

\bibitem[McDermid et~al.(2021)McDermid, Jia, Porter, and Habli]{mcdermid2021artificial}
John~A McDermid, Yan Jia, Zoe Porter, and Ibrahim Habli.
\newblock Artificial intelligence explainability: the technical and ethical dimensions.
\newblock \emph{Philosophical Transactions of the Royal Society A}, 379:\penalty0 20200363--20200363, 2021.

\bibitem[Tomsett et~al.(2020)Tomsett, Harborne, Chakraborty, Gurram, and Preece]{tomsett2020sanity}
Richard Tomsett, Dan Harborne, Supriyo Chakraborty, Prudhvi Gurram, and Alun Preece.
\newblock Sanity checks for saliency metrics.
\newblock In \emph{Proceedings of the AAAI conference on artificial intelligence}, volume~34, pages 6021--6029, 2020.

\bibitem[Zafar and Khan(2021)]{rehman2019dlime}
Muhammad~Rehman Zafar and Naimul Khan.
\newblock Deterministic local interpretable model-agnostic explanations for stable explainability.
\newblock \emph{Machine Learning and Knowledge Extraction}, 3\penalty0 (3):\penalty0 525--541, 2021.

\bibitem[Doshi-Velez and Kim(2018)]{doshi2017towards}
Finale Doshi-Velez and Been Kim.
\newblock Considerations for evaluation and generalization in interpretable machine learning.
\newblock \emph{Explainable and interpretable models in computer vision and machine learning}, 1:\penalty0 3--17, 2018.

\bibitem[Rosenfeld(2021)]{rosenfeld2021better}
Avi Rosenfeld.
\newblock Better metrics for evaluating explainable artificial intelligence.
\newblock In \emph{Proceedings of the 20th international conference on autonomous agents and multiagent systems}, pages 45--50, 2021.

\bibitem[Kadir et~al.(2023)Kadir, Mosavi, and Sonntag]{10297629}
Md~Abdul Kadir, Amir Mosavi, and Daniel Sonntag.
\newblock Evaluation metrics for {XAI}: A review, taxonomy, and practical applications.
\newblock In \emph{2023 IEEE 27th International Conference on Intelligent Engineering Systems (INES)}, pages 000111--000124, 2023.

\bibitem[Paszke et~al.(2019)Paszke, Gross, Massa, Lerer, Bradbury, Chanan, Killeen, Lin, Gimelshein, Antiga, Desmaison, Kopf, Yang, DeVito, Raison, Tejani, Chilamkurthy, Steiner, Fang, Bai, and Chintala]{NEURIPS2019_9015}
Adam Paszke, Sam Gross, Francisco Massa, Adam Lerer, James Bradbury, Gregory Chanan, Trevor Killeen, Zeming Lin, Natalia Gimelshein, Luca Antiga, Alban Desmaison, Andreas Kopf, Edward Yang, Zachary DeVito, Martin Raison, Alykhan Tejani, Sasank Chilamkurthy, Benoit Steiner, Lu~Fang, Junjie Bai, and Soumith Chintala.
\newblock Pytorch: An imperative style, high-performance deep learning library.
\newblock In \emph{Advances in Neural Information Processing Systems 32}, pages 8024--8035. Curran Associates, Inc., 2019.
\newblock URL \url{http://papers.neurips.cc/paper/9015-pytorch-an-imperative-style-high-performance-deep-learning-library.pdf}.

\bibitem[Zhang and Xu(2024)]{zhang2024implicit}
Zhongwang Zhang and Zhi-Qin~John Xu.
\newblock Implicit regularization of dropout.
\newblock \emph{IEEE Transactions on Pattern Analysis and Machine Intelligence}, 2024.

\bibitem[Kimanius et~al.(2024)Kimanius, Jamali, Wilkinson, L{\"o}vestam, Velazhahan, Nakane, and Scheres]{kimanius2024data}
Dari Kimanius, Kiarash Jamali, Max~E Wilkinson, Sofia L{\"o}vestam, Vaithish Velazhahan, Takanori Nakane, and Sjors~HW Scheres.
\newblock Data-driven regularization lowers the size barrier of cryo-em structure determination.
\newblock \emph{Nature Methods}, pages 1--6, 2024.

\bibitem[{De Mol} et~al.(2009){De Mol}, {De Vito}, and Rosasco]{DEMOL2009201}
Christine {De Mol}, Ernesto {De Vito}, and Lorenzo Rosasco.
\newblock Elastic-net regularization in learning theory.
\newblock \emph{Journal of Complexity}, 25\penalty0 (2):\penalty0 201--230, 2009.
\newblock ISSN 0885-064X.
\newblock \doi{https://doi.org/10.1016/j.jco.2009.01.002}.
\newblock URL \url{https://www.sciencedirect.com/science/article/pii/S0885064X0900003X}.

\bibitem[Ross et~al.(2017)Ross, Hughes, and Doshi-Velez]{ijcai2017p371}
Andrew~Slavin Ross, Michael~C. Hughes, and Finale Doshi-Velez.
\newblock Right for the right reasons: Training differentiable models by constraining their explanations.
\newblock In \emph{Proceedings of the Twenty-Sixth International Joint Conference on Artificial Intelligence, {IJCAI-17}}, pages 2662--2670, 2017.

\bibitem[Ismail et~al.(2021)Ismail, Corrada~Bravo, and Feizi]{ismail2021improving}
Aya~Abdelsalam Ismail, Hector Corrada~Bravo, and Soheil Feizi.
\newblock Improving deep learning interpretability by saliency guided training.
\newblock \emph{Advances in Neural Information Processing Systems}, 34:\penalty0 26726--26739, 2021.

\bibitem[Uddin et~al.(2020)Uddin, Monira, Shin, Chung, Bae, et~al.]{uddin2020saliencymix}
AFM Uddin, Mst Monira, Wheemyung Shin, TaeChoong Chung, Sung-Ho Bae, et~al.
\newblock Saliencymix: A saliency guided data augmentation strategy for better regularization.
\newblock \emph{arXiv preprint arXiv:2006.01791}, 2020.

\bibitem[Jeffreys(1946)]{jeffreys1946invariant}
Harold Jeffreys.
\newblock An invariant form for the prior probability in estimation problems.
\newblock \emph{Proceedings of the Royal Society of London. Series A. Mathematical and Physical Sciences}, 186\penalty0 (1007):\penalty0 453--461, 1946.

\bibitem[Mir{\'o}-Nicolau et~al.(2024)Mir{\'o}-Nicolau, Jaume-i Cap{\'o}, and Moy{\`a}-Alcover]{miro2023assessing}
Miquel Mir{\'o}-Nicolau, Antoni Jaume-i Cap{\'o}, and Gabriel Moy{\`a}-Alcover.
\newblock Assessing fidelity in xai post-hoc techniques: A comparative study with ground truth explanations datasets.
\newblock \emph{Artificial Intelligence}, 335:\penalty0 104179, 2024.

\bibitem[Zhu and Ogino(2019)]{zhu2019guideline}
Peifei Zhu and Masahiro Ogino.
\newblock Guideline-based additive explanation for computer-aided diagnosis of lung nodules.
\newblock In \emph{Interpretability of Machine Intelligence in Medical Image Computing and Multimodal Learning for Clinical Decision Support}, pages 39--47, Cham, 2019. Springer International Publishing.

\bibitem[Schallner et~al.(2019)Schallner, Rabold, Scholz, and Schmid]{schallner2019effect}
Ludwig Schallner, Johannes Rabold, Oliver Scholz, and Ute Schmid.
\newblock Effect of superpixel aggregation on explanations in lime--a case study with biological data.
\newblock In \emph{Joint European Conference on Machine Learning and Knowledge Discovery in Databases}, pages 147--158, 2019.

\bibitem[Yan et~al.(2022)Yan, Hou, Liu, Yan, Huang, and Wang]{yan2022towards}
Anli Yan, Ruitao Hou, Xiaozhang Liu, Hongyang Yan, Teng Huang, and Xianmin Wang.
\newblock Towards explainable model extraction attacks.
\newblock \emph{International Journal of Intelligent Systems}, 37\penalty0 (11):\penalty0 9936--9956, 2022.

\bibitem[Krizhevsky et~al.(2010)Krizhevsky, Nair, and Hinton]{cifar10}
Alex Krizhevsky, Vinod Nair, and Geoffrey Hinton.
\newblock {CIFAR}-10 (canadian institute for advanced research).
\newblock \emph{URL http://www. cs. toronto. edu/kriz/cifar. html}, 5\penalty0 (4):\penalty0 1, 2010.

\bibitem[Krizhevsky(2009)]{cifar100}
Alex Krizhevsky.
\newblock Learning multiple layers of features from tiny images.
\newblock \emph{URL http://www. cs. toronto. edu/kriz/cifar. html}, pages 32--33, 2009.
\newblock URL \url{https://www.cs.toronto.edu/~kriz/learning-features-2009-TR.pdf}.

\bibitem[Xiao et~al.(2017)Xiao, Rasul, and Vollgraf]{xiao_fashion-mnist_2017}
Han Xiao, Kashif Rasul, and Roland Vollgraf.
\newblock Fashion-{MNIST}: a novel image dataset for benchmarking machine learning algorithms.
\newblock \emph{arXiv preprint arXiv:1708.07747}, 2017.

\bibitem[Cohen et~al.(2017)Cohen, Afshar, Tapson, and Van~Schaik]{cohen2017emnist}
Gregory Cohen, Saeed Afshar, Jonathan Tapson, and Andre Van~Schaik.
\newblock {EMNIST}: Extending mnist to handwritten letters.
\newblock In \emph{2017 International Joint Conference on Neural Networks (IJCNN)}, pages 2921--2926, 2017.

\bibitem[Nilsback and Zisserman(2008)]{nilsback2008automated}
Maria-Elena Nilsback and Andrew Zisserman.
\newblock Automated flower classification over a large number of classes.
\newblock In \emph{2008 Sixth Indian conference on computer vision, graphics \& image processing}, pages 722--729, 2008.

\bibitem[Parkhi et~al.(2012)Parkhi, Vedaldi, Zisserman, and Jawahar]{parkhi2012oxford}
Omkar~M Parkhi, Andrea Vedaldi, Andrew Zisserman, and CV~Jawahar.
\newblock The oxford-{IIIT} pet dataset.
\newblock In \emph{Proceedings of the IEEE Conference on Computer Vision and Pattern Recognition}, 2012.

\bibitem[Deng et~al.(2009)Deng, Dong, Socher, Li, Li, and Fei-Fei]{deng2009imagenet}
Jia Deng, Wei Dong, Richard Socher, Li-Jia Li, Kai Li, and Li~Fei-Fei.
\newblock Imagenet: A large-scale hierarchical image database.
\newblock In \emph{2009 IEEE conference on computer vision and pattern recognition}, pages 248--255. Ieee, 2009.

\bibitem[Tan and Le(2019)]{tan2019efficientnet}
Mingxing Tan and Quoc Le.
\newblock Efficientnet: Rethinking model scaling for convolutional neural networks.
\newblock In \emph{International conference on machine learning}, pages 6105--6114, 2019.

\bibitem[Tan and Le(2021)]{tan2021efficientnetv2}
Mingxing Tan and Quoc Le.
\newblock Efficientnetv2: Smaller models and faster training.
\newblock In \emph{International conference on machine learning}, pages 10096--10106, 2021.

\bibitem[Kingma and Ba(2014)]{kingma2014adam}
Diederik~P Kingma and Jimmy Ba.
\newblock Adam: A method for stochastic optimization.
\newblock \emph{arXiv preprint arXiv:1412.6980}, 2014.

\bibitem[Carrasco et~al.(2017)Carrasco, Garc{\'\i}a, del Mar~Rueda, and Herrera]{carrasco2017rnpbst}
Jacinto Carrasco, Salvador Garc{\'\i}a, Mar{\'\i}a del Mar~Rueda, and Francisco Herrera.
\newblock rnpbst: An r package covering non-parametric and bayesian statistical tests.
\newblock In \emph{Hybrid Artificial Intelligent Systems: 12th International Conference, Proceedings of Machine Learning Research 2017}, pages 281--292. Springer, 2017.

\bibitem[Ghosal(1997)]{ghosal1997review}
Subhashis Ghosal.
\newblock A review of consistency and convergence of posterior distribution.
\newblock In \emph{Varanashi Symposium in Bayesian Inference, Banaras Hindu University}. Citeseer, 1997.

\end{thebibliography}



\end{document}